\pdfoutput=1
\documentclass[letterpaper]{article} 
\usepackage[preprint]{aaai2027}  
\usepackage[hyphens]{url}  
\usepackage{graphicx} 
\usepackage{natbib}  
\usepackage{caption} 
\usepackage{algorithm}
\usepackage{algorithmic}

\usepackage{newfloat}
\usepackage{listings}
\DeclareCaptionStyle{ruled}{labelfont=normalfont,labelsep=colon,strut=off} 
\floatstyle{ruled}
\newfloat{listing}{tb}{lst}{}
\floatname{listing}{Listing}

\usepackage{booktabs}
\usepackage{xcolor}   
\usepackage{colortbl} 

\newcommand{\sig}{\rlap{$^{*}$}}

\title{Rethinking Verbalized Confidence for LLM-as-a-Judge:\\A Compatibility Shift on Post-2025 Proprietary Models}
\author{Yu-Chung Hsiao}
\affiliations{Cisco Systems}

\begin{document}

\maketitle

\begin{abstract}
Verbalized confidence, long dismissed as overconfident, coarse, and prone to round-number clustering, is now the more robust soft-scoring mechanism for LLM-as-a-Judge on top-tier proprietary models. Across SummEval, AggreFact, and HelpSteer2, spanning up to 18 LLMs, we show that the standard advice to prefer log-probabilities no longer holds on post-2025 models, where verbalized confidence is the better signal. We call this a \emph{compatibility shift}. On top of a standard verbalized-confidence baseline, we introduce two new ingredients: an overconfidence advisory and self-debate. Together they improve calibration, score-distribution spread, and robustness to task subjectivity. We further observe a generation effect: post-2025 models accommodate these two additions with little balanced-accuracy cost, whereas pre-2025 models pay a measurable penalty. Compared with logprob-based G-Eval, verbalized confidence is the more subjectivity-robust soft signal on GPT-family top-tier releases. The shift is invisible under accuracy-only reporting. Rather than defaulting to hard predictions, we recommend broader use of soft scoring in LLM-as-a-Judge. More broadly, verbalized confidence has moved from a weaker substitute for logprobs to a practical soft-scoring mechanism for contemporary LLM judges.
\end{abstract}

\section{Introduction}
\label{sec:intro}

LLM-as-a-Judge protocols prompt a general-purpose LLM to score a candidate output against a quality criterion. A central methodological question has been which \emph{soft signal} the judge should report: token log-probabilities (logprobs) over the prediction tokens, or a verbalized confidence the judge produces in natural language. Pre-2025 evidence weighed heavily toward logprobs. G-Eval observed that LLM Likert scores cluster on a few integer values, weakening rank correlations with human ratings such as Kendall's $\tau$. G-Eval therefore computes a finer-grained expected score from the token log-probabilities \citep{liu2023geval}. \citet{wang2025judgment} further showed that the G-Eval style soft-scoring protocol beats greedy verbal extraction in pointwise, pairwise, and listwise settings. Large-scale benchmarks reported that LLMs are systematically overconfident when verbalizing \citep{xiong2024canllms}, and recent work documented that verbal scores cluster on a handful of round-number values, collapsing their discriminative power \citep{dai2026rescaling}. The dominant practical recommendation, therefore, has been ``prefer logprobs.''

Most of this evidence, however, comes from pre-2025 models, and we find that the comparison looks different on contemporary proprietary flagships (each vendor's top-tier model). As frontier vendors moved through post-2025 release waves, two changes pull in opposite directions. On the \emph{logprob} side, access has become less stable: closed-model APIs increasingly restrict or remove logprob access. On the \emph{verbal} side, recent mechanistic work on open-weight models suggests that a model's verbalized confidence is more than its token probabilities restated in words, and reflects a richer internal assessment that can carry information not captured by the log-probabilities \citep{kumaran2026howllms}. Although we cannot verify this mechanistically on closed-source flagships, whose internals and often log-probabilities are inaccessible, it motivates us to probe them externally with a behavioral question: can a prompt recipe make the verbalized signal informative enough to match or exceed logprob-based soft scoring?

This work tests the question empirically across an 18-model sweep spanning the GPT, Claude, and Gemini families, sampling both pre-2025 checkpoints and post-2025 flagships. We find that a well-designed verbalized-confidence recipe is now a practical soft signal for frontier judges, and that on the one family where a fair comparison with logprobs is still feasible (GPT), our prompt recipe overtakes logprob-based G-Eval, reversing the pre-2025 preference. We call this reversal a \textbf{compatibility shift}: the better prompt guidance depends on the model generation, so advice calibrated on one generation can fail on a newer one. We trace the shift to a \textbf{generation effect}: post-2025 flagship models are better able to follow richer judge prompts, an overconfidence advisory and a self-debate that weighs both stances in a single call, improving calibration while keeping their balanced accuracy. Pre-2025 models lose significant balanced accuracy under the same prompts. This shift becomes visible when we analyze the soft score's calibration and score-distribution spread. Hence we recommend broader use of soft scoring in LLM-as-a-Judge, reported with these measures rather than hard-prediction accuracy alone.

\paragraph{Main claims and contributions.}
\begin{itemize}
\item \textbf{A logprob-free verbalized-confidence recipe} (Method) that adds only two ingredients, an overconfidence advisory and self-debate, on top of the established rubric-anchored baseline, yielding probability scores that are better calibrated, broader in spread, and less coupled to task subjectivity than that baseline, which is what makes verbalized confidence a strong soft signal in the first place.
\item \textbf{A compatibility shift in the soft signal} (Compatibility Shift section). On GPT-family flagships, the only family that still returns logprobs (Claude never did, and Gemini dropped them at 3.1 Pro), verbalized confidence overtakes logprob-based G-Eval as the more subjectivity-robust signal and steadily improves its alignment with human ratings across GPT generations, while G-Eval's fluctuates.
\item \textbf{A generation-effect explanation for the shift} (same section). Post-2025 flagship models keep their prediction accuracy when we add the advisory and self-debate, whereas pre-2025 models lose a significant amount. This difference is why the recipe helps current models without the accuracy cost it imposes on older ones.
\end{itemize}

\paragraph{Scope.}
LLM-as-a-Judge is the main tool for \emph{reference-free} quality evaluation: it scores a candidate without a gold reference. We study its single-call, pointwise form, where one model call returns a binary judgment together with a soft confidence score. The single call is a scalability choice, and also a foundational one: multi-call and multi-agent judging build on the same per-call judgment and confidence signal included in this study. Besides, evaluation trades per-example resources against data size, and one call per example keeps cost low to evaluate at the larger scale that gives stable, representative estimates. We study two kinds of judging target: objective factual faithfulness, through 9 tasks from AggreFact, and more subjective quality, through 3 tasks from SummEval (consistency, coherence, and relevance) and the HelpSteer2 benchmark, which themselves vary in subjectivity. Tasks with a checkable answer, such as coding, mathematics, and many agent objectives, are served by verifiable, reference-based evaluation aimed at correctness and fall outside our scope, together with pairwise and listwise judging and multi-call multi-agent debate. We center on contemporary closed-source flagships from the GPT, Claude, and Gemini families, with smaller and open-weight models as generation-effect controls.

\section{Method}
\label{sec:method}

\subsection{Proposed Soft-Scoring Protocols}
\label{sec:method-protocols}

\begin{figure}[t]
\centering
\includegraphics[width=0.85\linewidth]{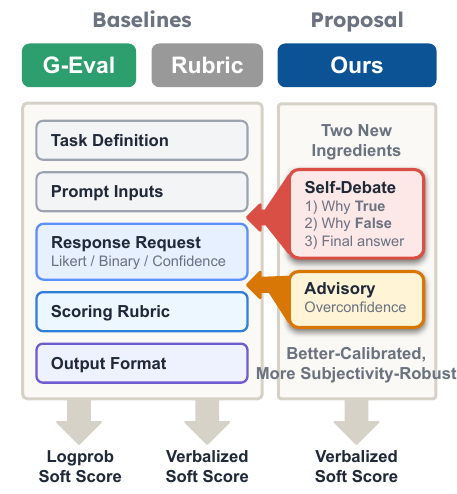}
\caption{Prompt protocols compared in this paper. G-Eval (logprob baseline) reads a \emph{logprob soft score} from token log-probabilities, whereas the verbalized protocols state the score in natural language, producing a \emph{verbalized soft score} together with a hard prediction (True/False). Rubric is the verbalized baseline. \textbf{Ours} adds two ingredients on top of Rubric, an overconfidence \emph{advisory} and a \emph{self-debate} step (argue True, argue False, then conclude), which give better-calibrated and more subjectivity-robust verbalized soft scores than Rubric. All protocols share the remaining blocks: task definition, prompt inputs, response request, scoring rubric, and output format. The Prompt design section details how each block is instantiated for logprob versus verbalized scoring. Ours-FreeReason, an ablation that replaces the self-debate step with free-form reasoning, is defined there but not drawn here.}
\label{fig:prompt-protocols}
\end{figure}

We construct probability scores from a pure verbal signal. The goal is to extract a soft score that (1) discriminates strongly between True and False on a given quality criterion and (2) is well calibrated. We constrain the use of LLMs to:
\begin{enumerate}
\item A single LLM call per judging. This excludes multi-agent setups.
\item Pointwise binary classification. Traditional machine-learning classifiers routinely provide graded quality scores, for example between $0$ and $1.0$. LLMs can be turned into ad-hoc binary classifiers through prompting, but they usually lack the granularity of the soft scoring that traditional models provide.
\end{enumerate}
Logprob-based methods such as G-Eval \citep{liu2023geval} and the judgment-distribution line \citep{wang2025judgment} are established alternatives, but logprob access has become less reliable across closed-model APIs (see Supplementary Material). We therefore focus on a logprob-free protocol.

We propose two Verbalized Confidence Protocols, \textbf{Ours-FreeReason} and \textbf{Ours}, with two baselines: the rubric-only \textbf{Rubric} variant and logprob-based \textbf{G-Eval} as comparison anchors (Figure~\ref{fig:prompt-protocols}).

\paragraph{Prompt components.}
Two components distinguish Ours from the Rubric baseline, and because their exact wording matters we reproduce it here (SummEval Relevance instantiation; full templates in the Supplementary Material). The \textbf{advisory} warns the judge against overconfidence, e.g.:
\begin{quote}\footnotesize\itshape
You have been consistently overconfident in past evaluations...
\end{quote}
\noindent The \textbf{self-debate request} asks the judge to argue both answers before deciding, e.g.:
\begin{quote}\footnotesize\itshape
Emulate a debate in three steps. (1)~True because: argue for ``True'' \dots (2)~False because: argue for ``False'' \dots (3)~Explanation: based on both arguments, determine your final answer.
\end{quote}

\paragraph{Prompt design.}
We view judge prompts as compositions of reusable components. The three shared blocks (response request, scoring rubric, output format) are instantiated to match each protocol's output type. G-Eval requests a Likert rating, so its scoring rubric anchors the Likert scale and its output is a single Likert value from which the logprob soft score is read. The verbalized protocols instead request a binary prediction together with a 0--100 confidence score, so their scoring rubric is a \emph{confidence rubric} that fixes how confidence is stated and their output lists the answer followed by the confidence. We define three verbalized protocols in this study:
\begin{itemize}
\item \textbf{Rubric (baseline):} a logprob-free verbalized-confidence protocol that asks the model for a binary judgment and an explicit confidence score under a confidence rubric.
\item \textbf{Ours-FreeReason (ablation):} the Rubric baseline augmented with advisory text discouraging overconfidence and with {\it free-form reasoning} before the prediction. The model is asked for explicit, unconstrained reasoning but is not required to consider both stances. 
\item \textbf{Ours:} the full recipe, which augments Rubric with advisory text and the self-debate request as above.
\end{itemize}
Relative to the Rubric baseline, only two components change across our protocols: the overconfidence advisory and the reasoning block (free-form for Ours-FreeReason, self-debate for Ours). The confidence rubric, output order, and evidence-quoting sub-directive are held constant across all three, so comparing Ours-FreeReason with Ours isolates the effect of self-debate versus free-form reasoning. The verbatim prompt templates are given in the Supplementary Material due to their length.

\subsection{Metrics}
\label{sec:method-metrics}

Table~\ref{tab:metric-guide} defines the six measurements and points to where each is used in the experiments. We report them together rather than in isolation, because no single one reveals the compatibility shift and only their combination does, a point we return to in the Discussion.

\begin{table}[t]
\centering\footnotesize
\setlength{\tabcolsep}{4pt}
\begin{tabular}{@{}l p{3.0cm} c l@{}}
\toprule
Metric & Measures & Dir. & Appears in \\
\midrule
Prediction BA & Balanced accuracy of the emitted binary prediction & $\uparrow$ & Tab.~\ref{tab:aggrefact-results},\,\ref{tab:ablation-summary} \\
Oracle BA & Best balanced accuracy from an oracle threshold on the confidence & $\uparrow$ & Fig.~\ref{fig:oracle-prediction-gap} \\
OP Gap & Oracle BA minus Prediction BA: the soft signal the hard prediction discards & $\downarrow$ & Fig.~\ref{fig:oracle-prediction-gap} \\
Debate Stress & Prediction-BA drop from self-debate vs.\ free-form reasoning & $\downarrow$ & Fig.~\ref{fig:oracle-prediction-gap} \\
AECE & Adaptive (equal-mass) expected calibration error & $\downarrow$ & Tab.~\ref{tab:aggrefact-results},\,\ref{tab:ablation-summary} \\
Spread & Bhattacharyya overlap of the confidence histogram with uniform & $\uparrow$ & Tab.~\ref{tab:aggrefact-results},\,\ref{tab:ablation-summary} \\
\bottomrule
\end{tabular}
\caption{The six metrics: what each measures, its better direction ($\uparrow$/$\downarrow$), and where it appears. AECE is expected calibration error with adaptive equal-mass bins \citep{guo2017calibration,nixon2019measuring}. Spread's formula is given in the text.}
\label{tab:metric-guide}
\end{table}

Two definitions need a further word. Oracle BA optimizes the decision threshold directly on the evaluation set, so we treat it as the ideal prediction accuracy recoverable from the confidence, and the (non-negative) Oracle--Prediction Gap (OP Gap) is the portion of that signal the emitted prediction leaves unused. Spread is the Bhattacharyya coefficient between the confidence histogram $\hat q = (\hat q_1,\dots,\hat q_B)$ over $B$ equal-width bins on $[0,1]$ and a uniform reference,
\[
\mathrm{Spread}(\hat q)=\frac{1}{\sqrt{B}}\sum_{b=1}^{B}\sqrt{\hat q_b},
\]
which we prefer over entropy because we want closeness to the uniform distribution, not generic dispersion.

\section{Experimental Setup}
\label{sec:setup}

\subsection{Tasks}
\label{sec:setup-tasks}

We use three benchmarks. SummEval and HelpSteer2 both probe robustness to task subjectivity, with HelpSteer2 serving as an independent replication of the SummEval result, while AggreFact probes binary faithfulness under prompt stress.

\textbf{SummEval} \citep{fabbri2021summeval} is a graded subjective evaluation: 100 documents each paired with 17 machine-generated summaries and expert 5-point Likert ratings. We use three dimensions, each a separate judging task. \textbf{Consistency} asks whether the summary remains faithful to the source; \textbf{Coherence} asks whether the summary reads as a well-formed paragraph; and \textbf{Relevance} asks whether the summary captures the important content while avoiding redundant or tangential details.

Inter-annotator agreement (IAA) for each SummEval dimension is Krippendorff's $\alpha$ on the expert 5-point ratings, with lower $\alpha$ indicating greater task subjectivity: Consistency $\alpha=0.80$ (least subjective), Coherence $\alpha=0.55$ (more subjective), and Relevance $\alpha=0.40$ (highly subjective). We use IAA as a per-task subjectivity covariate in the slope analyses below.

\textbf{HelpSteer2} \citep{wang2024helpsteer2} is a modern LLM-output-quality benchmark with roughly 1{,}000 crowd annotators and 3--5 raters per item over about 21k prompt-response pairs. Its crowd annotation and task diversity differ sharply from SummEval's three-expert panel, so we use it as an independent replication of the subjectivity result rather than a pooled extension. We sample 800 items. Because HelpSteer2's task diversity precludes a single per-task agreement value, we measure subjectivity per example from rater disagreement across its 3--5 ratings, the analogue of SummEval's IAA.

\textbf{AggreFact} \citep{tang-etal-2023-understanding} is a binary factual-faithfulness suite spanning multiple problem regimes: RAG hallucination detection (RAGTruth), retrieval-augmented claim verification (ClaimVerify), expert-QA grounding (ExpertQA), long-form QA grounding (Lfqa), multi-step reasoning verification (Reveal), dialog and meeting grounding (TofuEval-MediaS, TofuEval-MeetB), post-hoc LLM fact-checking (FactCheck-GPT), and news-summary faithfulness (AggreFact-CNN). We use these 9 tasks with 300 examples each (2{,}700 total). Task selection and preprocessing are described in the Supplementary Material.

\subsection{Models}
\label{sec:setup-models}

Our full sweep covers 18 LLMs spanning the GPT, Claude, and Gemini families plus open-weight checkpoints. Three cohorts are used. The \textbf{AggreFact cohort} is a 10-model selection of closed-source flagships, grouped by training generation into \emph{pre-2025} (GPT-4o, GPT-4.1, o1, sonnet-4, gemini-2.5-pro) and \emph{post-2025} (GPT-5.2, GPT-5.4, sonnet-4.5, sonnet-4.6, gemini-3.1-pro). The \textbf{SummEval cohort} extends this to 16 models by adding 5 open-weight models (gpt-oss-20b, gpt-oss-120b, llama-3.3-70b, mistral-large-3, gemma-4-26b) and 3 smaller closed models (gpt-4o-mini, gpt-4.1-mini, gpt-4.1-nano), in order to probe whether subjectivity robustness extends beyond flagship scale. The \textbf{HelpSteer2 cohort} is a 9-model flagship set spanning 4 vendors and both eras: \emph{pre-2025} (GPT-4o, GPT-4.1, sonnet-4, gemini-2.5-pro, mistral-large-3) and \emph{post-2025} (GPT-5.2, GPT-5.4, sonnet-4.5, gemini-3.1-pro).

For the GPT-family G-Eval comparison, which requires logprob access, we use the OpenAI flagships that return logprobs (gpt-4o, gpt-4.1, gpt-5.2, gpt-5.4). GPT is the only frontier family that still returns logprobs across several releases, so the comparison is restricted to it by necessity, not by choice: Claude has never returned logprobs, and Gemini returned them at 2.5 Pro but withdrew them at 3.1 Pro (Supplementary Material). This withdrawal is itself part of the picture, since it makes a logprob-free protocol the forward-compatible option on the newest models.

\section{Prompt Effect: The Recipe Improves Soft-Scoring Quality}
\label{sec:prompt-effect}

Across all three datasets, our recipe improves soft-scoring quality. On AggreFact, the recipe lowers calibration error and broadens the score distribution. On SummEval and HelpSteer2, it further makes the probability signal more robust to task subjectivity.

\subsection{Calibration Error and Score Spread Improve}
\label{sec:prompt-aggrefact}

\begin{figure}[t]
\centering
\includegraphics[width=\linewidth]{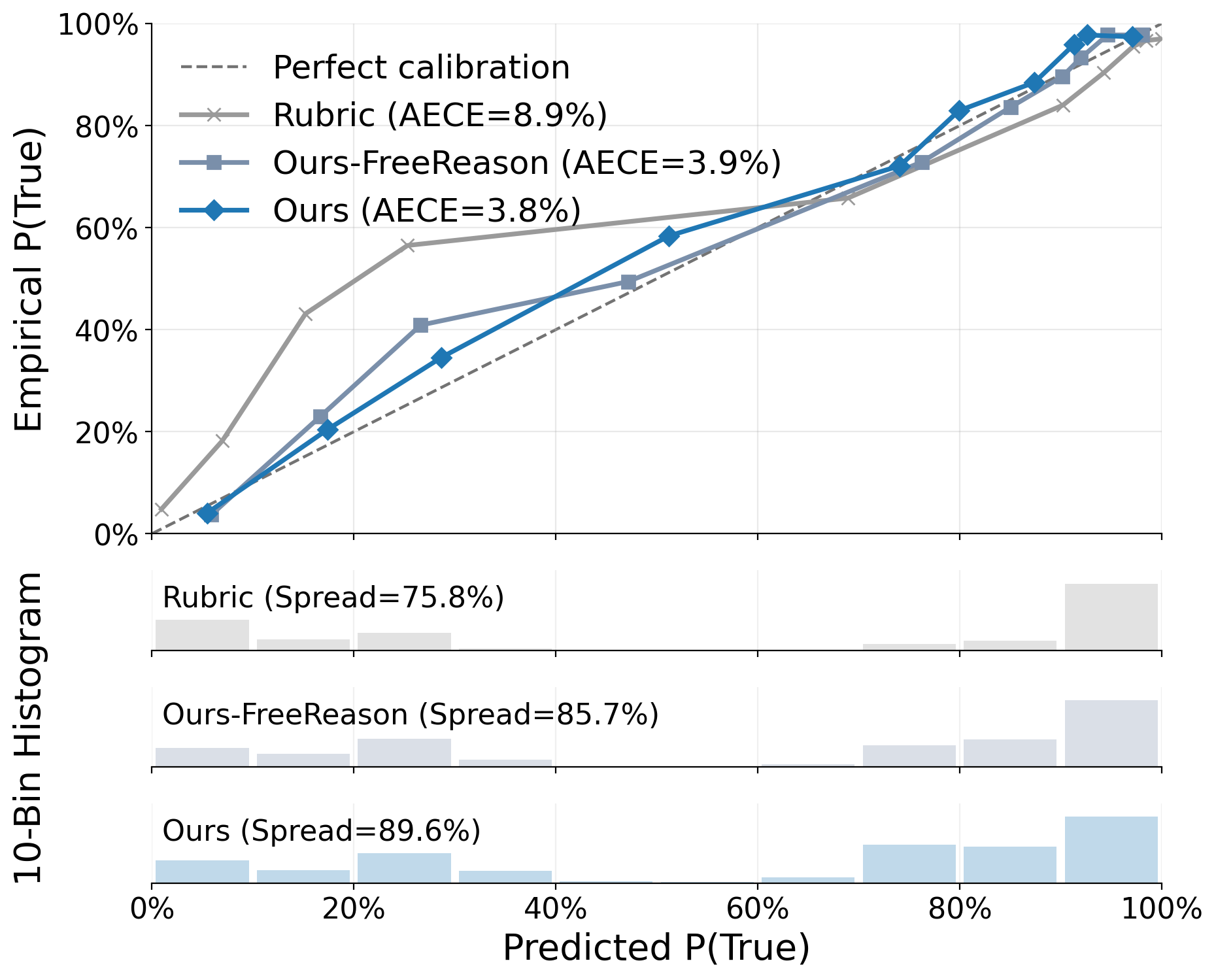}
\caption{Calibration curves and probability histograms for Sonnet 4.5 on AggreFact. The top panel plots overall calibration curves using 10 equal-mass bins; points with fewer than 27 examples are omitted. The lower panels show 10-bin histograms of predicted probabilities. Compared with Rubric, our prompts substantially reduce calibration error (AECE from 8.9\% to 3.8\%) and make the probability distribution more evenly spread (Spread from 75.8\% to 89.6\%).}
\label{fig:sonnet45-calibration}
\end{figure}

Relative to Rubric, Ours-FreeReason sharply reduces calibration error, and Ours preserves this gain while further broadening the score distribution. On Sonnet 4.5 for AggreFact, our prompts make the predicted probabilities less collapsed and more evenly distributed across the probability range (Figure~\ref{fig:sonnet45-calibration}). The rubric baseline is especially overconfident in low-probability regions. Our prompts tame this overconfident pattern, especially on the false side, bringing the calibration curve closer to the diagonal while maintaining broader use of the probability scale. The full per-model calibration-error and spread results appear in Table~\ref{tab:aggrefact-results}.

\subsection{Ours: Soft Signal More Robust to Subjectivity}
\label{sec:prompt-summeval}

\begin{figure}[t]
\centering
\includegraphics[width=0.85\linewidth]{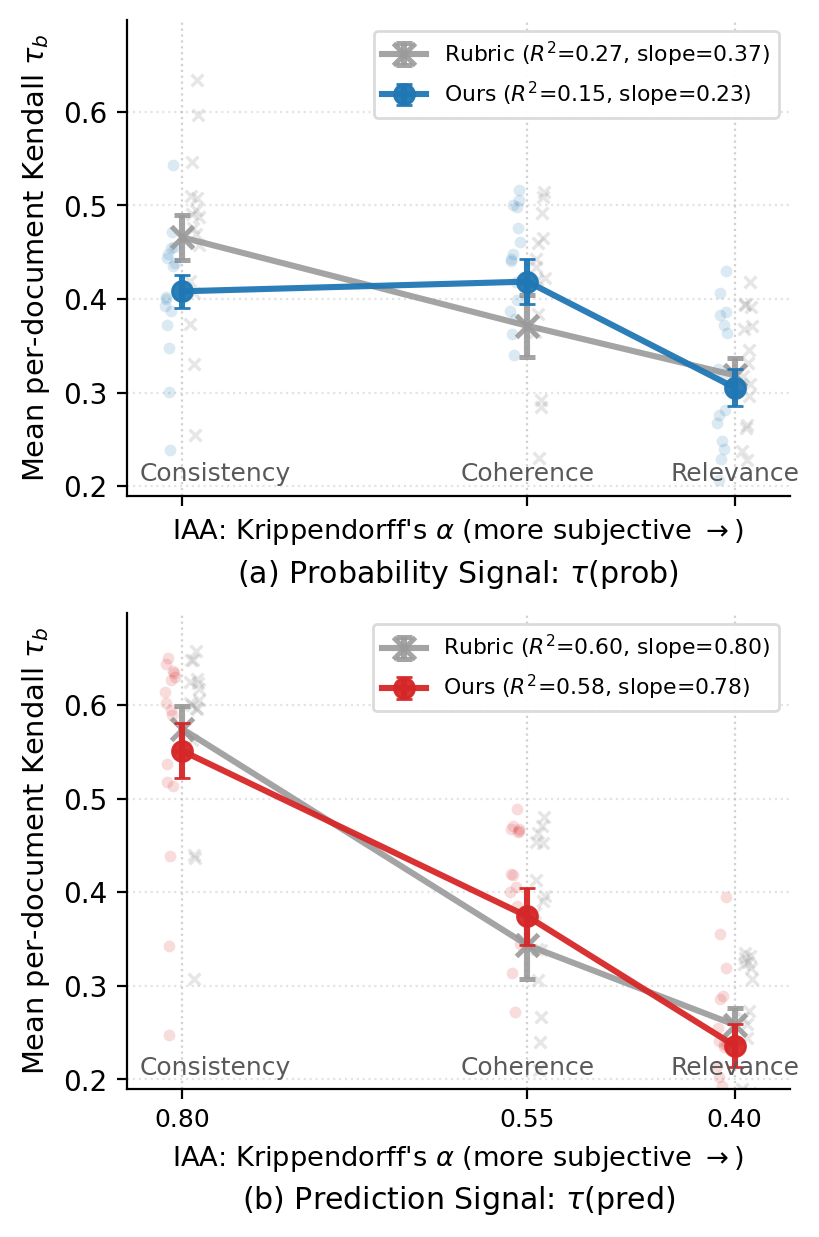}
\caption{Probability signals in panel~(a) are more robust to task subjectivity than prediction signals in panel~(b). Smaller regression slopes and lower $R^2$ indicate weaker coupling to inter-annotator agreement (IAA) and thus a less confounded capacity measure. Faint markers show per-model values, jittered along the $x$-axis for visualization only. Relative to the baseline prompt, our prompt significantly flattens $\tau(\texttt{prob})$ ($\Delta\mathrm{slope}=-0.143$, one-sided $p<10^{-4}$), whereas the slope difference between the two prompts is not significant for $\tau(\texttt{pred})$ (two-sided $p=0.46$).}
\label{fig:tau-subjectivity}
\end{figure}

Turning to SummEval, we computed Kendall's $\tau_b$ between the judgments of our 16 tested LLMs and the expert-annotated Likert ratings, separately for binary predictions and probability scores for each prompt. We then averaged these per-model correlations across the models, denoting the resulting means $\tau(\texttt{prob})$ and $\tau(\texttt{pred})$, plotted in Figure~\ref{fig:tau-subjectivity} against IAA on the $x$-axis. We compared two prompt configurations, Rubric and Ours.

Our prompt's overconfidence advisory and self-debate make the probability signal significantly more robust to task subjectivity than the Rubric baseline, flattening its $\tau(\texttt{prob})$ slope by $\Delta\mathrm{slope}=-0.143$ and lowering $R^2$ (one-sided $p<10^{-4}$, panel~a). This gain is clearest in the soft-score signal: the probability slopes in panel~(a) are flatter than the binary-prediction slopes in panel~(b) (one-sided $p<10^{-5}$), and the dimension-level IAA slope does not separate the two prompts on $\tau(\texttt{pred})$ (two-sided $p=0.46$). The slope is estimated over 48 model-and-task points ($16$ models $\times$ $3$ dimensions), shown as the per-model markers in Figure~\ref{fig:tau-subjectivity}. Its confidence intervals and $p$-values come from a paired model-cluster bootstrap that resamples models with replacement while preserving each model's task points.
The SummEval annotation process recruited the same expert pool for all dimensions through the same training, so we treat the IAA gradient as genuine subjectivity rather than noise.

\paragraph{Ours' robustness advantage replicates independently.}
Krippendorff's $\alpha$ is interpretable only within a single study's design \citep{krippendorff2011computing}, so $\alpha$ points cannot be pooled across datasets with different annotator pools and rubrics. We therefore corroborate the subjectivity-robustness conclusion with a complementary per-example rater-disagreement measure that does not require dimension homogeneity, and we replicate on HelpSteer2 (described in the Tasks section). Treated as independent replications rather than a pooled fit, all analyses agree that Ours is more robust to subjectivity than Rubric: Ours' slope is significantly flatter on both datasets and both signals (SummEval $\Delta\mathrm{slope}=-0.024$ for $\tau(\texttt{prob})$ with $p=0.018$ and $-0.058$ for $\tau(\texttt{pred})$ with $p=0.004$; HelpSteer2 $\Delta\mathrm{slope}=-0.023$ with $p=0.004$ and $-0.036$ with $p<10^{-4}$), and its $R^2$ is lower in every case. This finer measure resolves the recipe's robustness gain on $\tau(\texttt{pred})$ too, which the dimension-level IAA slope could not. The HelpSteer2 replication spans an entirely independent annotation infrastructure. Full tables are in the Supplementary Material.

\section{Compatibility Shift and Generation Effect}
\label{sec:model-effect}

This section presents the shift and then the generation effect behind it. First, on SummEval, GPT-family flagships realign from logprob-based G-Eval toward verbalized confidence, the compatibility shift itself. We then trace this shift to a generation effect: on AggreFact, post-2025 models keep their prediction accuracy when the advisory and self-debate are added, shrinking the Oracle-Prediction Gap and Debate Stress toward zero, whereas older models incur a prediction-quality penalty. 

\subsection{Verbalized Confidence Overtakes Logprob G-Eval on GPT-Family Flagships}
\label{sec:model-summeval}

We compare four judging protocols on the GPT flagships that return logprobs (gpt-4o, gpt-4.1, gpt-5.2, gpt-5.4): G-Eval, Rubric, Ours-FreeReason, and Ours. For each task and protocol, we compute mean per-document Kendall $\tau_b$ between model scores and expert ratings across candidate summaries, and quantify subjectivity robustness as the slope of $\tau(\texttt{prob})$ against task IAA $\alpha$. Lower slopes indicate less sensitivity to task subjectivity.

\begin{figure}[t]
\centering
\includegraphics[width=0.9\linewidth]{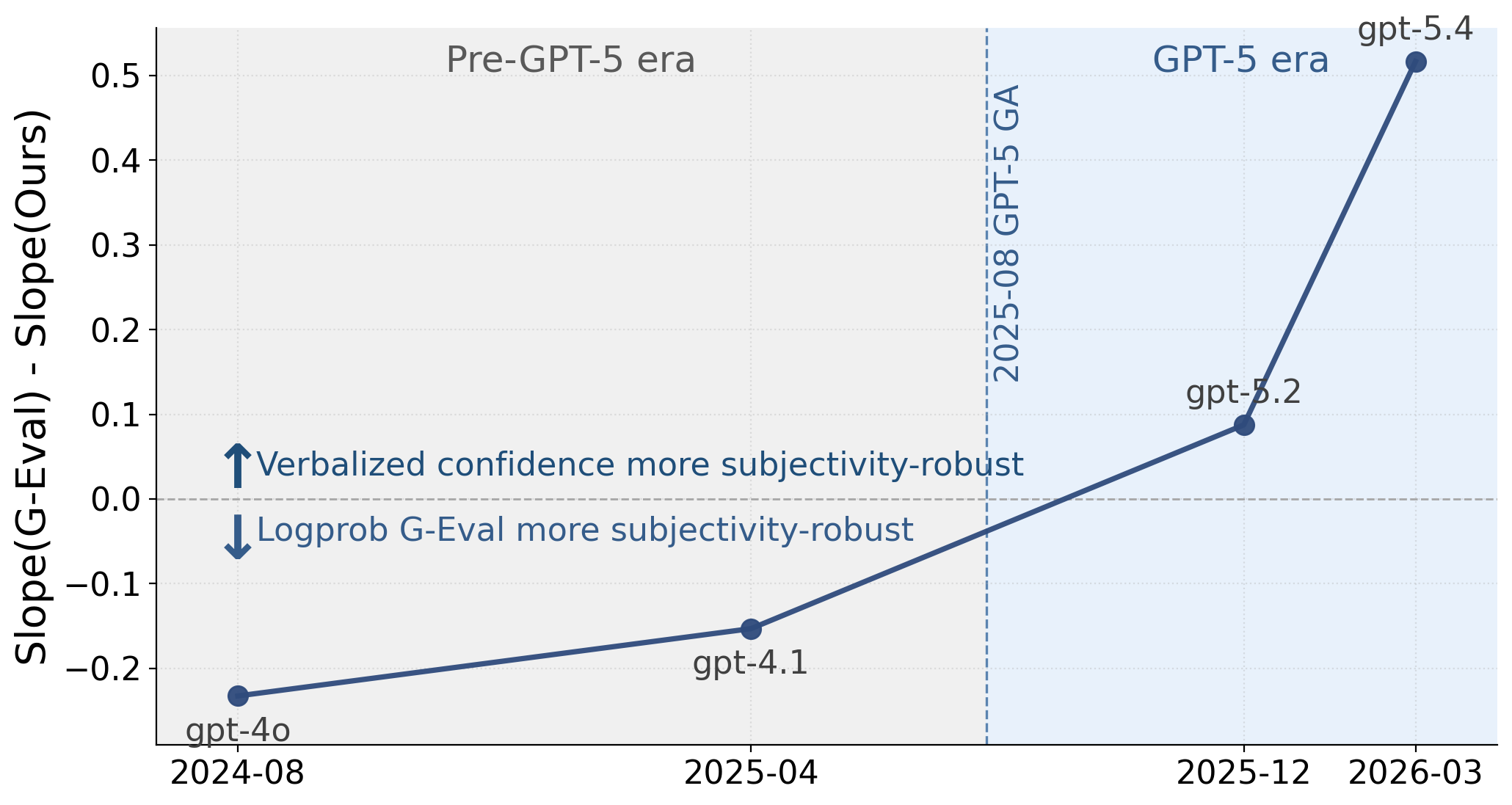}\\
{\small\textbf{(a)}~Verbalized confidence's \emph{subjectivity-robustness} advantage over G-Eval.}\\[4pt]
\includegraphics[width=0.9\linewidth]{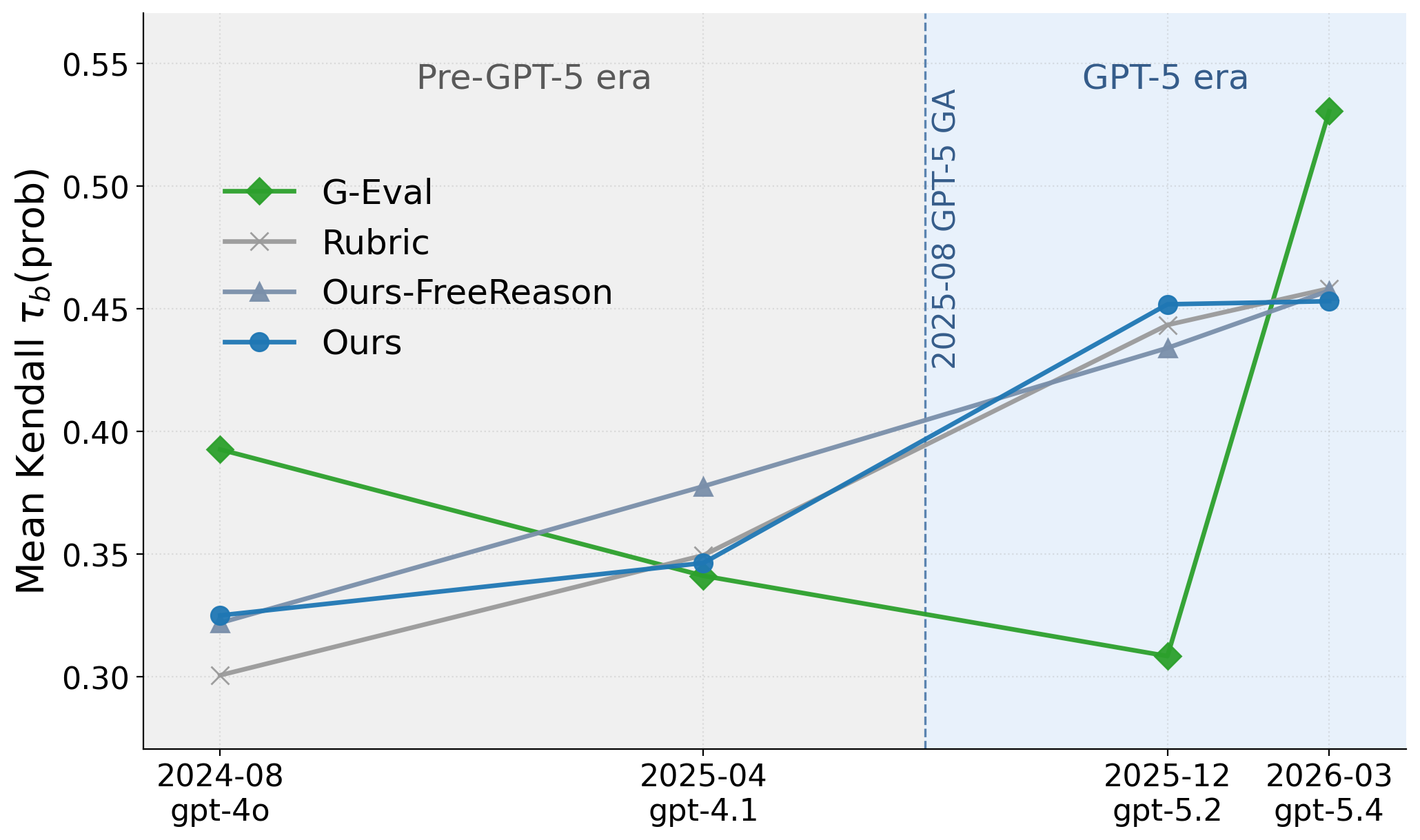}\\
{\small\textbf{(b)}~Mean Kendall $\tau_b$ between the soft score and human ratings.}
\caption{Compatibility shift across GPT releases on SummEval. The advantage in (a) changes sign from negative (pre-GPT-5) to positive (GPT-5 era), so GPT-5 era models favor verbalized confidence over logprob-based G-Eval for subjectivity robustness. In (b), the verbalized protocols (Rubric, Ours-FreeReason, Ours) improve steadily across GPT generations, whereas logprob-based G-Eval fluctuates. Together, verbalized confidence overtakes logprob-based G-Eval as the more compatible soft signal for contemporary GPT-family flagships.}
\label{fig:flagships}
\end{figure}

Both views in Figure~\ref{fig:flagships} favor newer GPT generations: verbalized confidence gains a subjectivity-robustness advantage over G-Eval in the GPT-5 era (a), and its judging capacity improves steadily across the family while G-Eval's does not (b). Verbalized confidence is therefore better aligned with current GPT-family models than logprob-based soft scoring, one practical reason to recommend Ours as the forward-compatible LLM-as-a-Judge protocol.

\subsection{Generation Effect: Newer Models Keep Their Accuracy With the Recipe, Older Models Don't}
\label{sec:model-aggrefact}

We now turn to the generation effect behind the shift, and ask whether the model's emitted binary prediction is as informative as the confidence signal verbalized in the same response. AggreFact is a binary faithfulness task, so this setting lets us directly compare the quality of the hard prediction against the quality of a thresholded confidence score. We compare three verbal protocols, Rubric, Ours-FreeReason, and Ours, using Prediction BA, Oracle BA, the Oracle-Prediction Gap, and Debate Stress.

Table~\ref{tab:aggrefact-results} reports Prediction BA and calibration-quality gains by family. Two patterns emerge. Across nearly all models, Ours yields large calibration gains, especially in score-distribution spread. The prediction-quality tradeoff is generation-dependent: older models incur significant BA drops while post-2025 flagships do not, and some even improve. These results are stable under a maximally-disjoint 300-per-task resample (Supplementary Material).

\begin{table*}[t]
\centering
\small
\begin{tabular}{lrrrrrrrrr}
\toprule
& \multicolumn{3}{c}{BA $\uparrow$} & \multicolumn{3}{c}{AECE $\downarrow$} & \multicolumn{3}{c}{Spread $\uparrow$} \\
\cmidrule(lr){2-4} \cmidrule(lr){5-7} \cmidrule(lr){8-10}
Model & Rubric & Ours & $\Delta\uparrow$ & Rubric & Ours & $\Delta\downarrow$ & Rubric & Ours & $\Delta\uparrow$ \\
\midrule
gpt-4o & 81.3\% & 80.2\% & $-1.1$ & 8.0\% & 4.1\% & $-3.9$\sig & 66.0\% & 88.8\% & $+22.8$\sig \\
o1 & 80.6\% & 79.2\% & $-1.4$\sig & 11.0\% & 5.8\% & $-5.2$\sig & 48.7\% & 71.9\% & $+23.2$\sig \\
gpt-4.1 & 81.7\% & 79.3\% & $-2.4$\sig & 11.2\% & 4.4\% & $-6.8$\sig & 53.5\% & 86.9\% & $+33.4$\sig \\
gpt-5.2$^\dagger$ & 82.4\% & 83.0\% & $+0.6$ & 9.7\% & 7.6\% & $-2.2$\sig & 66.9\% & 83.6\% & $+16.6$\sig \\
gpt-5.4$^\dagger$ & 81.3\% & 81.9\% & $+0.6$ & 12.7\% & 9.1\% & $-3.6$\sig & 67.2\% & 81.0\% & $+13.7$\sig \\
\midrule
sonnet-4 & 81.3\% & 80.1\% & $-1.2$\sig & 8.4\% & 5.9\% & $-2.4$\sig & 71.3\% & 81.5\% & $+10.2$\sig \\
sonnet-4.5$^\dagger$ & 80.9\% & 82.0\% & $+1.1$ & 9.1\% & 3.3\% & $-5.8$\sig & 75.8\% & 89.6\% & $+13.9$\sig \\
sonnet-4.6$^\dagger$ & 81.5\% & 82.7\% & $+1.2$\sig & 10.2\% & 8.1\% & $-2.2$\sig & 81.9\% & 86.4\% & $+4.4$\sig \\
\midrule
gemini-2.5-pro & 80.7\% & 78.0\% & $-2.7$\sig & 17.7\% & 13.5\% & $-4.2$\sig & 45.8\% & 66.3\% & $+20.5$\sig \\
gemini-3.1-pro$^\dagger$ & 83.3\% & 83.0\% & $-0.3$ & 13.1\% & 11.2\% & $-1.8$\sig & 52.5\% & 64.2\% & $+11.6$\sig \\
\bottomrule
\end{tabular}
\caption{AggreFact prediction accuracy and probability-signal quality under Rubric and Ours. Asterisks (*) denote statistical significance under paired bootstrapping. Daggers ($^\dagger$) denote post-2025 flagship models. Ours consistently reduces calibration error and broadens use of the probability scale, while prediction-accuracy costs are concentrated in older models.}
\label{tab:aggrefact-results}
\end{table*}

\begin{figure}[t]
\centering
\includegraphics[width=\linewidth]{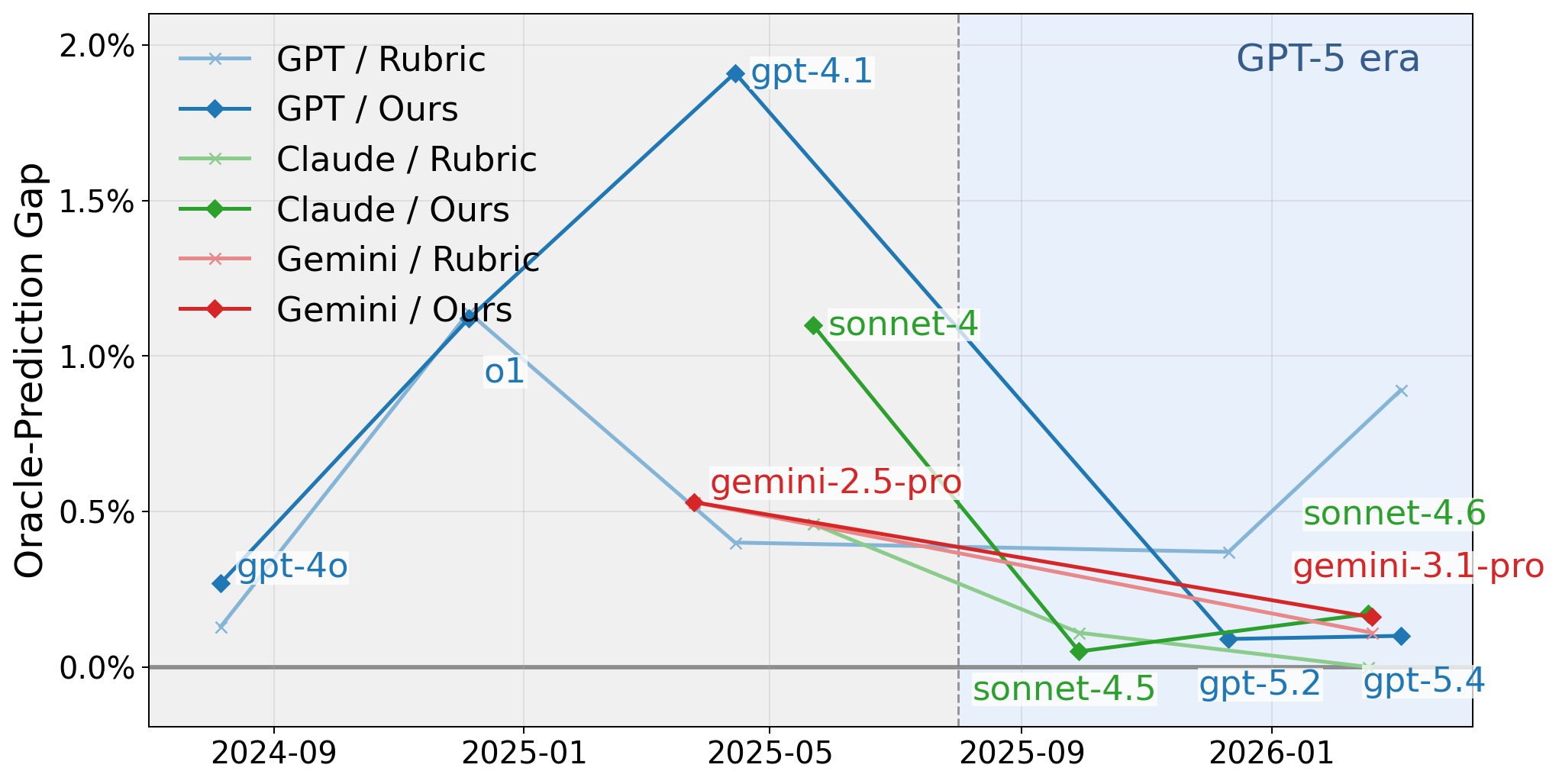}\\
{\footnotesize\textbf{(a)} Oracle-Prediction Gap}\\[3pt]
\includegraphics[width=\linewidth]{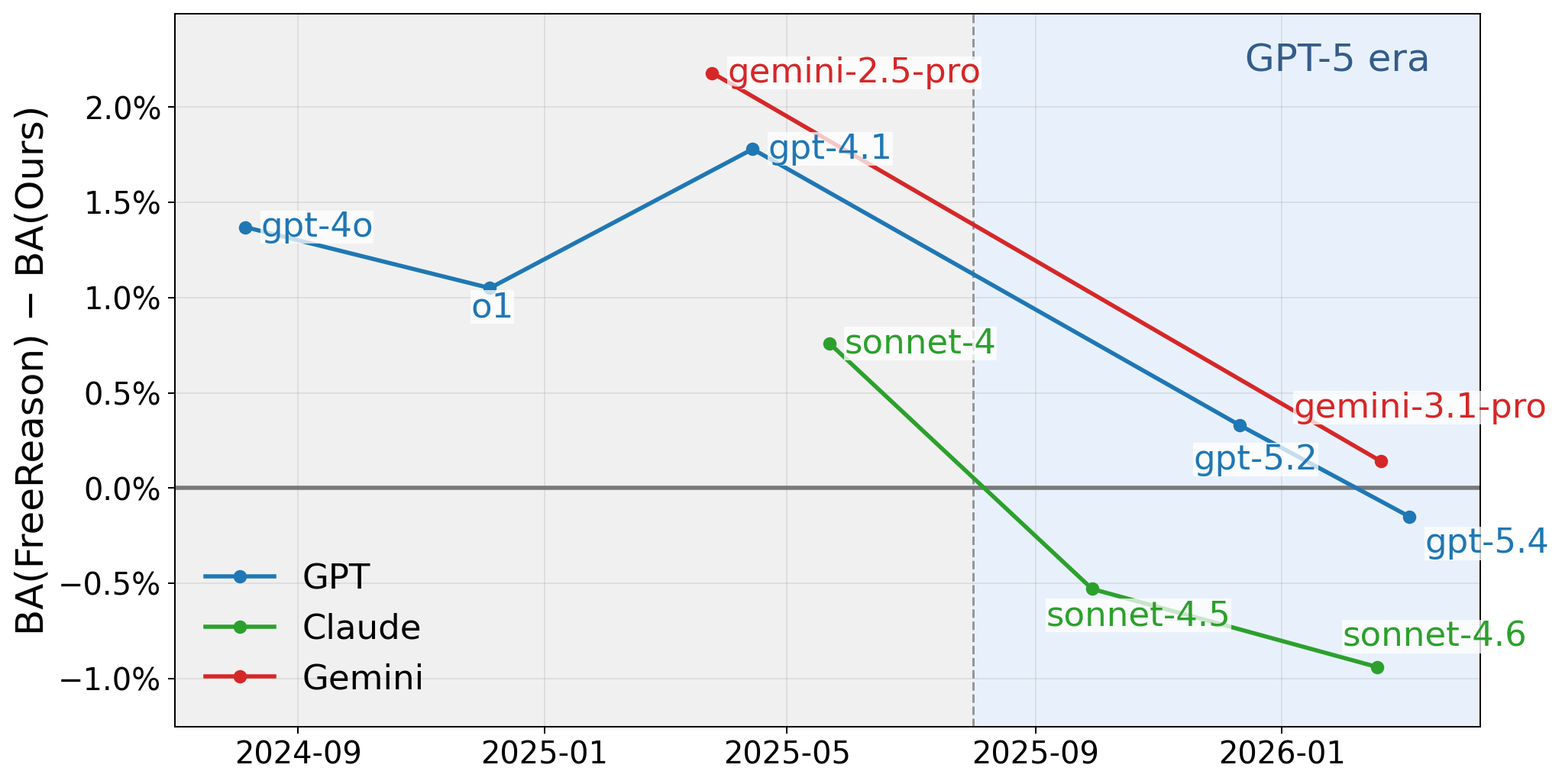}\\
{\footnotesize\textbf{(b)} Debate Stress}
\caption{Two prompt-stress measures over model generations, lower is better. \textbf{(a)}~Oracle-Prediction Gap: a smaller gap means the binary prediction better captures the faithfulness signal in the confidence. \textbf{(b)}~Debate Stress: the prediction-BA change from self-debate, with zero meaning full robustness and negative values a gain. Both shrink toward zero on post-2025 flagships.}
\label{fig:oracle-prediction-gap}
\end{figure}

The generation effect appears most clearly in the two prompt-stress measures. Under Ours, newer flagships shrink the Oracle-Prediction Gap toward zero after mid-2025 (Figure~\ref{fig:oracle-prediction-gap}a), and Debate Stress follows the same trajectory (Figure~\ref{fig:oracle-prediction-gap}b): older GPT and Gemini models show substantial positive stress while post-2025 models move close to zero, and later Claude models even gain prediction BA from debate. The full per-family comparison is in the Supplementary Material.

\subsection{Which Component Drives Which Effect}
\label{sec:ablation}

To isolate the contribution of each moving component, we ran a 10-model, 5-configuration sweep on AggreFact that adds the advisory and the reasoning block independently. Starting from Rubric, we add the advisory (+Adv), then either free-form reasoning (+FR) or self-debate (+SD) on top, and separately apply self-debate on its own against Rubric. Table~\ref{tab:ablation-summary} summarizes each configuration split by model era, and the full per-model grid across all three metrics is in the Supplementary Material. Three findings emerge. First, the \textbf{advisory is the primary calibration lever}: it significantly reduces AECE and broadens Spread in every cohort, whereas self-debate on its own barely moves AECE and contributes only to Spread. Second, \textbf{free-form reasoning erodes the advisory's spread gain while self-debate preserves it}: relative to +Adv, free-form significantly reduces Spread in both eras, whereas self-debate preserves it and augments it on pre-2025 models. Third, \textbf{the prediction-accuracy effect is a clean generation split}: the full Adv+SD recipe significantly lowers pre-2025 cohort BA but significantly raises it on post-2025 models, and both the advisory and self-debate contribute this pre-2025-only cost individually. The advisory and self-debate contributions are approximately additive on calibration, and their joint condition is where the compatibility shift is most visible.

\begin{table}[!t]
\centering\footnotesize
\setlength{\tabcolsep}{4pt}
\begin{tabular}{@{}lccclrrr}
\toprule
Models & Adv & FR & SD & Base & $\Delta$BA & $\Delta$AECE & $\Delta$Spread \\
\midrule
Pre  & $\surd$ &         &         & Rubric & $-0.7$\sig & $-4.5$\sig & $+18.9$\sig \\
Post & $\surd$ &         &         & Rubric & $+0.0$ & $-3.6$\sig & $+11.6$\sig \\
\cmidrule(lr){1-8}
Pre  & $\surd$ & $\surd$ &         & +Adv & $+0.4$\sig & $+0.1$ & $-2.5$\sig \\
Post & $\surd$ & $\surd$ &         & +Adv & $+0.4$ & $+0.4$ & $-3.1$\sig \\
\cmidrule(lr){1-8}
Pre  & $\surd$ &         & $\surd$ & \textbf{+Adv} & {\boldmath$-1.0$\sig} & $-0.0$ & $+3.0$\sig \\
Post & $\surd$ &         & $\surd$ & \textbf{+Adv} & {\boldmath$+0.6$\sig} & $+0.5$ & $+0.5$ \\
\cmidrule(lr){1-8}
Pre  & $\surd$ &         & $\surd$ & \textbf{Rubric} & {\boldmath$-1.8$\sig} & $-4.5$\sig & $+22.0$\sig \\
Post & $\surd$ &         & $\surd$ & \textbf{Rubric} & {\boldmath$+0.7$\sig} & $-3.1$\sig & $+12.1$\sig \\
\cmidrule(lr){1-8}
Pre  &         &         & $\surd$ & Rubric & $-0.8$\sig & $-0.0$ & $+6.1$\sig \\
Post &         &         & $\surd$ & Rubric & $+0.3$ & $+0.2$ & $+2.1$\sig \\
\bottomrule
\end{tabular}
\caption{Component-isolation ablation on AggreFact. Adv: overconfidence advisory; FR: free-form reasoning; SD: self-debate. Pre/Post: pre-2025 and post-2025 flagship models. Base: the configuration each $\Delta$ is measured against (checkmarks mark active blocks). Each cell is the mean $\Delta$ (percentage points) over the row's group; * marks a 95\% confidence interval excluding zero under a paired model-cluster bootstrap ($B=20{,}000$). Bold marks the full-recipe (Adv+SD, i.e.\ Ours) $\Delta$Prediction-BA cells, where the generation effect is starkest: {\boldmath\bfseries pre-2025 models lose balanced accuracy ($-1.8$~pp) while post-2025 models gain ($+0.7$~pp)}. Per-model detail is in the Supplementary Material.}
\label{tab:ablation-summary}
\end{table}

\subsection{Robustness Across Tasks}
\label{sec:subtask}

The recipe's benefit holds per task, not only in aggregate. We score the full recipe on every model-and-task combination, 90 in all, and bootstrap each separately (Supplementary Material). The calibration benefit is broad: 67 of the 90 (model, task) comparisons favor Ours (binomial one-sided $p=1.9\times10^{-6}$), none favors Rubric at 95\% confidence, and it appears on all 9 tasks rather than concentrating on a few. The prediction-accuracy cost is again generation-dependent, with significant $\Delta$BA drops concentrated on the pre-2025 cohort (Fisher one-sided $p=0.013$).

\section{Discussion}
\label{sec:discussion}

The prompt and model effects are complementary. Our recipe improves the probability signal across settings: on SummEval and HelpSteer2 it makes $\tau(\texttt{prob})$ more robust to task subjectivity, and on AggreFact it cuts calibration error while broadening the score distribution. Post-2025 flagships then convert this richer signal into stable predictions, whereas older models often pay a prediction-accuracy penalty for the same calibration gains. This is why accuracy-only reporting misses the shift: $\Delta$Prediction BA looks flat on newer models, while $\Delta$AECE alone hides whether the prediction still preserves the latent signal.

For benchmark practice, we therefore recommend reporting the soft-score measures alongside prediction accuracy, and, as logprob access grows less reliable across production APIs, adopting a logprob-free verbalized-confidence protocol as the more portable soft-scoring interface.

\section{Related Work}
\label{sec:related-work}

\paragraph{Soft scoring and verbalized confidence.}
LLM-as-judge is well studied \citep{zheng2023judging,liu2023geval}, with AggreFact \citep{tang2023aggrefact} and SummEval \citep{fabbri2021summeval} as standard faithfulness testbeds. G-Eval \citep{liu2023geval} and later judgment-distribution methods \citep{wang2025judgment} use token logprobs to recover soft signal beyond greedy outputs. Prior evidence on verbalized confidence is mixed: it can be useful on QA correctness \citep{kadavath2022know,tian2023just}, but is often overconfident or collapsed onto round numbers \citep{xiong2024canllms,dai2026rescaling}. Mechanistic work suggests verbal confidence can reflect internal answer-quality assessments beyond a logprob readout \citep{kumaran2026howllms}. We show this signal is practically useful on contemporary closed-source flagships.

\paragraph{Self-debate and calibration.}
Unlike multi-call debate or aggregation systems \citep{du2023improving,liang2023encouraging,khan2024debating,jiang2023llmblender,wang2024moa}, our self-debate is a single-call prompt constraint. For soft-score quality, we pair adaptive ECE \citep{guo2017calibration,nixon2019measuring} with a Bhattacharyya spread against a uniform reference \citep{chao2019quantifying,gregorius2021concept,pierson2021usinggeometryrankevenness}, since ECE alone can miss the distributional collapse common in verbalized scores \citep{dai2026rescaling}.

\section{Conclusion}
\label{sec:conclusion}

The pre-2025 preference for logprob-based soft scoring in LLM-as-a-Judge no longer holds uniformly. Across our three benchmarks, our logprob-free verbalized-confidence recipe improves calibration, score spread, and robustness to task subjectivity, while post-2025 flagships keep their prediction accuracy and pre-2025 models lose it, the generation effect behind the compatibility shift. In practice, report hard predictions and soft-score measures together, and treat prompt-model compatibility as a moving target.

\bibliography{custom}

\clearpage
\appendix
\section*{Supplementary Material}

\noindent This appendix collects the supplementary material referred to from the main text. Each section is self-contained:
\begin{itemize}
\item \textbf{Section A}: logprob availability across recent APIs.
\item \textbf{Sections B, C}: the full subjectivity-robustness statistics, the bootstrap protocol (B) and the per-example replication on SummEval and HelpSteer2 (C).
\item \textbf{Sections D, E}: the per-model component ablation (D) and the per-task replication on AggreFact (E).
\item \textbf{Sections F, G}: data processing (deterministic sampling, task selection, and sizes) in F, and its stability under an independent resample in G.
\item \textbf{Section H}: per-family AggreFact metric trends across model generations.
\item \textbf{Section I}: the verbatim prompt templates.
\end{itemize}

\section{A. Logprob Access Is No Longer a Stable Primitive}
\label{app:logprob}

A clear trend emerges across recent closed-model APIs: token logprobs are becoming less uniformly available as a general-purpose evaluation signal. Earlier flagship models return them cleanly, which makes logprob-based soft scoring straightforward. Newer model families increasingly restrict that path: support becomes conditional on model variant or serving configuration rather than being a stable default capability. Even where logprobs remain accessible, the maximum number of top tokens returned has been reduced. The GPT family, for example, dropped from 20 in GPT-4o/4.1 to 5 in GPT-5.2/5.4 (Table~\ref{tab:logprob-availability}). Support can also be removed outright across a family's releases: Google's Gemini exposed logprobs at 2.5 Pro but withdrew them starting from 3.0 Pro, and Anthropic's Claude has never returned them. This makes logprob-based judge protocols harder to apply consistently across modern frontier models, and in several families impossible on the newest release. Conversely, logprob-free confidence extraction remains broadly portable, which is why it is the more practical soft-signal interface for cross-model evaluation.

\begin{table*}[t]
\centering\small
\setlength{\tabcolsep}{5pt}
\begin{tabular}{@{}lccr@{}}
\toprule
Model & Release & \texttt{logprobs} available? & Max \texttt{top\_logprobs} \\
\midrule
\rowcolor{green!6}  GPT-4o          & 2024-08 & Yes               & 20 \\
\rowcolor{green!6}  o1              & 2024-12 & No                &  \\
\rowcolor{green!6}  GPT-4.1         & 2025-04 & Yes               & 20 \\
\rowcolor{green!6}  GPT-5           & 2025-08 & No                &  \\
\rowcolor{green!6}  GPT-5.2         & 2025-12 & Config dependent  & 5  \\
\rowcolor{green!6}  GPT-5.4         & 2026-03 & Config dependent  & 5  \\
\rowcolor{cyan!8}   Gemini 2.5 Pro  & 2025-06 & Yes               & 20 \\
\rowcolor{cyan!8}   Gemini 3.0 Pro  & 2025-11 & No                &  \\
\rowcolor{cyan!8}   Gemini 3.1 Pro  & 2026-02 & No                &  \\
\rowcolor{orange!10}All Claude Models & ---   & No                &  \\
\bottomrule
\end{tabular}
\caption{Logprob availability across recent closed-model APIs. Row shading indicates vendor: \colorbox{green!6}{OpenAI}, \colorbox{cyan!8}{Google}, \colorbox{orange!10}{Anthropic}. Within the OpenAI family, availability has narrowed over time. Google withdrew logprob support entirely starting from Gemini 3.0 Pro. Anthropic's Claude has never supported logprobs.}
\label{tab:logprob-availability}
\end{table*}

\section{B. Bootstrap Inference for Subjectivity-Sensitivity Slopes}
\label{app:bootstrap}

Slope differences are estimated by simple linear regression of $\tau_b$ against the subjectivity covariate, with confidence intervals and $p$-values from a paired model-cluster bootstrap ($B=100{,}000$) that resamples models with replacement while preserving each model's task or bin points. The resampling unit is the \emph{model} in all cases, so the bootstrap evaluates whether Ours' slope is reliably flatter than Rubric's across the model cohort. Two variants of this procedure are used:

\begin{itemize}
\item \textbf{SummEval (dimension-level IAA):} 16 models, regressing $\tau_b$ against Krippendorff's $\alpha$ computed at the dimension level (3 non-fluency dimensions). Table~\ref{tab:summeval-subjectivity} reports these results.
\item \textbf{SummEval and HelpSteer2 (per-example disagreement):} 16 models for SummEval and 9 flagship models for HelpSteer2, regressing $\tau_b$ against bin-mean rater disagreement, i.e.\ the std of per-annotator ratings per example averaged within each equal-mass bin (full definition in Section~C). Table~\ref{tab:subjectivity-extended} reports these results.
\end{itemize}

\begin{table*}[ht]
\centering\footnotesize
\setlength{\tabcolsep}{4pt}
\begin{tabular}{@{}llcccrrcc@{}}
\toprule
Signal & Prompt & Cons. & Coh. & Rel. & Slope$\downarrow$ & $R^2\downarrow$ & $\Delta$slope & $p$ \\
\midrule
$\tau(\texttt{pred})$ & Rubric & 0.58 & 0.34 & 0.26 & 0.80 & 0.61 & & \\
   & Ours & 0.55 & 0.37 & 0.24 & 0.78 & 0.59 & $-0.025$ & $0.46$ \\
\midrule
$\tau(\texttt{prob})$ & Rubric & 0.47 & 0.37 & 0.32 & 0.37 & 0.27 & & \\
   & Ours & 0.41 & 0.42 & 0.30 & \textbf{0.23} & \textbf{0.15} & $\mathbf{-0.143}$ & $\mathbf{<10^{-4}}$ \\
\bottomrule
\end{tabular}
\caption{SummEval dimension-level subjectivity-robustness results. Per-dimension mean Kendall $\tau_b$ (Cons.~= Consistency, Coh.~= Coherence, Rel.~= Relevance), slope and $R^2$ from regression on IAA (Krippendorff's $\alpha$), and bootstrap slope-difference tests ($\Delta$slope = Ours minus Rubric; $B=100{,}000$ paired model-cluster bootstrap). }
\label{tab:summeval-subjectivity}
\end{table*}

\section{C. Per-Example Subjectivity Robustness}
\label{app:subjectivity-extended}

The main paper reports a complementary per-example rater-disagreement measure and an independent HelpSteer2 replication. This section gives the full protocol and results. Table~\ref{tab:summeval-subjectivity} gives the per-model, per-dimension $\tau_b$ values underlying the SummEval dimension-level slope analysis.

\paragraph{Per-example subjectivity measure.}
We measure subjectivity per example rather than only per dimension: we bin examples by the standard deviation of their per-annotator ratings and regress $\tau$ against bin-mean disagreement. This gives finer resolution than dimension-level IAA, and because it does not require dimension homogeneity it applies uniformly to SummEval and to datasets such as HelpSteer2, whose prompt-response pairs span coding, creative writing, science, and professional communication rather than a single domain. It also keeps the measurement comparable within each study: Krippendorff's $\alpha$ is interpretable only within a single study's design \citep{krippendorff2011computing}, since rater training, rubric anchors, and pool size are part of the measurement scale (an $\alpha=0.5$ on SummEval's expert panel is not the same construct as $\alpha=0.5$ on another dataset's crowd pool), so we report each dataset on its own scale rather than pooling $\alpha$ across studies.

\paragraph{HelpSteer2 setup.}
HelpSteer2 \citep{wang2024helpsteer2} is a modern LLM-output-quality benchmark with roughly 1{,}000 crowd annotators, about 21k prompt-response pairs, and 3--5 raters per item on a 0--4 Likert scale. We downsample 800 items via the same md5-hash convention as the AggreFact draw (Section~F), and judge with 9 flagship models across 4 vendors and two eras (pre-2025: gpt-4.1, gpt-4o, sonnet-4, gemini-2.5-pro, mistral-large-3; post-2025: gpt-5.2, gpt-5.4, sonnet-4.5, gemini-3.1-pro). Because HelpSteer2 items span a wide task mix within each dimension, per-dimension $\alpha$ collapses genuinely different subjectivity regimes into one number. We therefore report a per-example rater-disagreement measure, computing $\alpha$ on rater ratings binarized at $\geq 3$. 
This qualitative boundary follows the HelpSteer2 annotation rubric \citep{wang2024helpsteer2}.
Specifically, the Correctness dimension requires a score of $\geq 3$ to indicate no hallucinations or misleading information, whereas $< 3$ indicates a mix of correct and incorrect content; likewise, $\geq 3$ on Coherence requires the response to be mostly clear with only minor issues, whereas $< 3$ indicates inconsistencies or contradictions. This cut also corresponds to the True/False boundary used in our judge prompts for HelpSteer2.
\begin{table*}[t]
\centering\footnotesize
\setlength{\tabcolsep}{6pt}
\begin{tabular}{@{}llrrrrrr@{}}
\toprule
Dataset & Signal & slope(Rubric) & slope(Ours) & $R^2$(Rubric) & $R^2$(Ours) & $\Delta$slope & $p$-value \\
\midrule
SummEval   & $\tau(\texttt{prob})$ & $+0.092$ & $+0.067$ & 0.11 & 0.08 & $-0.024$ & $0.018$ \\
SummEval   & $\tau(\texttt{pred})$ & $+0.237$ & $+0.171$ & 0.25 & 0.19 & $-0.058$ & $0.004$ \\
HelpSteer2 & $\tau(\texttt{prob})$ & $+0.280$ & $+0.258$ & 0.85 & 0.80 & $-0.023$ & $0.004$ \\
HelpSteer2 & $\tau(\texttt{pred})$ & $+0.292$ & $+0.256$ & 0.85 & 0.74 & $-0.036$ & $<10^{-4}$ \\
\bottomrule
\end{tabular}
\caption{Per-example rater-disagreement slope differences on SummEval and HelpSteer2. Ours' slope is significantly flatter than Rubric's on both datasets and both signals, and Ours' $R^2$ is lower in every case. Because Krippendorff's $\alpha$ is not comparable across studies with different annotator pools and rubrics, the two datasets are reported separately, yet reach the same statistical conclusion.}
\label{tab:subjectivity-extended}
\end{table*}

\section{D. Prompt Component Ablation \\ (10 Models $\times$ 5 Configs)}
\label{app:ablation}

Relative to Rubric, our recipe (Ours) adds two components: the overconfidence \textbf{advisory} and \textbf{self-debate}. For a thorough comparison, we include Ours-FreeReason, which substitutes self-debate with a free-form chain-of-thought reasoning block. Other prompt components, such as the task description, confidence rubric, and output order, are held constant across all protocols. Paired-bootstrap comparisons use the same recipe as Table~2 in the main paper (1{,}000 replicates, $\alpha=0.05$, shared example keys).

\textbf{The advisory is the primary calibration lever}: adding it reduces AECE and broadens Spread on all 10 models (10/10 significant on both), and removing it from Ours reverses both on all 10 (Tables~\ref{tab:abl-aece} and~\ref{tab:abl-spread}). \textbf{Self-debate adds Spread on top of the advisory} on 6/10 models, whereas free-form reasoning is roughly neutral or slightly negative on Spread. Prediction BA results are in Table~\ref{tab:abl-ba}. The generation effect is driven by advisory and self-debate acting together.

\begin{table*}[t]
\centering\footnotesize
\setlength{\tabcolsep}{4pt}
\begin{tabular}{@{}llrr>{\boldmath\bfseries\columncolor{gray!12}}rr>{\boldmath\bfseries\columncolor{gray!12}}rr>{\boldmath\bfseries\columncolor{gray!12}}rr>{\boldmath\bfseries\columncolor{gray!12}}r}
\toprule
Generation & Model & Rubric & +Adv & $\Delta$Adv & +FR & $\Delta$FR & +SD & $\Delta$SD & ($-$Adv) & $\Delta(-$Adv$)$ \\
\midrule
\rowcolor{green!6}   Pre-2025 & gpt-4o         & 81.3\% & 80.9\% & $-0.4$ & 81.6\% & $+0.7$ & 80.2\% & $-0.7$ & 79.4\% & $-0.9$ \\
\rowcolor{green!6}            & o1             & 80.6\% & 80.1\% & $-0.5$ & 80.3\% & $+0.1$ & 79.2\% & $-0.9$\sig           & 80.5\% & $+1.3$\sig           \\
\rowcolor{green!6}            & gpt-4.1        & 81.7\% & 81.2\% & $-0.5$ & 81.0\% & $-0.1$ & 79.3\% & $-1.9$\sig           & 80.8\% & $+1.6$\sig           \\
\rowcolor{orange!10}          & sonnet-4       & 81.3\% & 80.1\% & $-1.2$ & 80.9\% & $+0.8$ & 80.1\% & $+0.0$ & 81.6\% & $+1.5$\sig           \\
\rowcolor{cyan!8}             & gemini-2.5-pro & 80.7\% & 79.7\% & $-1.0$\sig           & 80.2\% & $+0.5$ & 78.0\% & $-1.7$\sig           & 79.5\% & $+1.5$\sig           \\
\midrule
\rowcolor{green!6}   Post-2025 & gpt-5.2        & 82.4\% & 82.7\% & $+0.3$ & 83.3\% & $+0.7$ & 83.0\% & $+0.3$ & 82.6\% & $-0.5$ \\
\rowcolor{green!6}             & gpt-5.4        & 81.3\% & 80.6\% & $-0.7$ & 81.8\% & $+1.1$\sig           & 81.9\% & $+1.3$\sig           & 81.3\% & $-0.6$ \\
\rowcolor{orange!10}           & sonnet-4.5     & 80.9\% & 80.9\% & $+0.0$ & 81.5\% & $+0.5$ & 82.0\% & $+1.1$ & 82.0\% & $-0.0$ \\
\rowcolor{orange!10}           & sonnet-4.6     & 81.5\% & 82.0\% & $+0.5$ & 81.8\% & $-0.2$ & 82.7\% & $+0.7$ & 82.0\% & $-0.7$\sig           \\
\rowcolor{cyan!8}            & gemini-3.1-pro & 83.3\% & 83.4\% & $+0.1$ & 83.2\% & $-0.2$ & 83.0\% & $-0.3$ & 83.0\% & $-0.0$ \\
\bottomrule
\end{tabular}
\caption{Prediction BA (\%; \textbf{higher is better}) per model across five configurations on AggreFact. \textbf{+Adv}: Rubric plus advisory. \textbf{+FR}: Rubric plus advisory plus free-form chain-of-thought (i.e., Ours-FreeReason). \textbf{+SD}: Rubric plus advisory plus self-debate (i.e., Ours). \textbf{($-$Adv)}: Ours with the advisory removed. Bold $\Delta$ columns show marginal changes: $\Delta$Adv is the gain from Rubric $\to$ +Adv, $\Delta$FR and $\Delta$SD are the gains from +Adv $\to$ +FR and +Adv $\to$ +SD, and $\Delta(-$Adv$)$ is the gain from +SD $\to$ ($-$Adv). Asterisks (*) denote bootstrap-significant deltas ($\alpha = 0.05$). Three findings stand out. (1) The advisory alone is largely neutral on prediction BA (1/10 significant). (2) Adding self-debate on top of the advisory drops BA significantly on 3/5 pre-2025 models, while post-2025 models are unaffected or gain (gpt-5.4: $+1.3$\sig); free-form reasoning shows no consistent effect in either generation. (3) Removing the advisory from Ours recovers BA on 4 pre-2025 models and hurts 1 post-2025 model, confirming that advisory and self-debate together drive the pre-2025 compatibility cost.}
\label{tab:abl-ba}
\end{table*}

\begin{table*}[t]
\centering\footnotesize
\setlength{\tabcolsep}{4pt}
\begin{tabular}{@{}llrr>{\boldmath\bfseries\columncolor{gray!12}}rr>{\boldmath\bfseries\columncolor{gray!12}}rr>{\boldmath\bfseries\columncolor{gray!12}}rr>{\boldmath\bfseries\columncolor{gray!12}}r}
\toprule
Generation & Model & Rubric & +Adv & $\Delta$Adv & +FR & $\Delta$FR & +SD & $\Delta$SD & ($-$Adv) & $\Delta(-$Adv$)$ \\
\midrule
\rowcolor{green!6}   Pre-2025 & gpt-4o         & 8.0\%  & 4.8\%  & $-3.2$\sig           & 5.1\%  & $+0.3$ & 4.1\%  & $-0.7$ & 10.9\% & $+6.8$\sig           \\
\rowcolor{green!6}            & o1             & 11.0\% & 7.4\%  & $-3.6$\sig           & 7.3\%  & $-0.1$ & 5.8\%  & $-1.6$ & 10.5\% & $+4.7$\sig           \\
\rowcolor{green!6}            & gpt-4.1        & 11.2\% & 4.0\%  & $-7.1$\sig           & 5.4\%  & $+1.4$ & 4.4\%  & $+0.3$ & 11.0\% & $+6.6$\sig           \\
\rowcolor{orange!10}          & sonnet-4       & 8.4\%  & 4.8\%  & $-3.6$\sig           & 4.8\%  & $-0.0$ & 5.9\%  & $+1.2$ & 8.2\%  & $+2.2$\sig           \\
\rowcolor{cyan!8}             & gemini-2.5-pro & 17.7\% & 12.9\% & $-4.8$\sig           & 11.8\% & $-1.1$ & 13.5\% & $+0.7$ & 15.7\% & $+2.1$\sig           \\
\midrule
\rowcolor{green!6}   Post-2025 & gpt-5.2       & 9.7\%  & 5.8\%  & $-3.9$\sig           & 7.3\%  & $+1.5$ & 7.6\%  & $+1.8$\sig           & 10.6\% & $+3.1$\sig           \\
\rowcolor{green!6}             & gpt-5.4       & 12.7\% & 7.1\%  & $-5.6$\sig           & 8.4\%  & $+1.2$ & 9.1\%  & $+2.0$\sig           & 11.6\% & $+2.5$\sig           \\
\rowcolor{orange!10}           & sonnet-4.5    & 9.1\%  & 4.2\%  & $-4.9$\sig           & 3.4\%  & $-0.8$ & 3.3\%  & $-0.9$ & 7.9\%  & $+4.6$\sig           \\
\rowcolor{orange!10}           & sonnet-4.6    & 10.2\% & 7.5\%  & $-2.7$\sig           & 6.8\%  & $-0.7$ & 8.1\%  & $+0.5$ & 10.2\% & $+2.2$\sig           \\
\rowcolor{cyan!8}              & gemini-3.1-pro & 13.1\% & 11.9\% & $-1.1$\sig          & 12.6\% & $+0.7$\sig           & 11.2\% & $-0.7$ & 15.3\% & $+4.0$\sig           \\
\bottomrule
\end{tabular}
\caption{AECE (\%; \textbf{lower is better}) per model across five configurations on AggreFact. Column definitions follow Table~\ref{tab:abl-ba}. Asterisks (*) denote bootstrap-significant deltas ($\alpha=0.05$). Two findings stand out. (1) The advisory reduces AECE on all 10 models (10/10 significant), confirming it as the primary calibration driver; neither self-debate nor free-form reasoning moves AECE significantly on pre-2025 models, and self-debate modestly worsens it on the two post-2025 GPT models. (2) Removing the advisory from Ours reverses the AECE gain on all 10 models (10/10 significant).}
\label{tab:abl-aece}
\end{table*}

\begin{table*}[t]
\centering\footnotesize
\setlength{\tabcolsep}{4pt}
\begin{tabular}{@{}llrr>{\boldmath\bfseries\columncolor{gray!12}}rr>{\boldmath\bfseries\columncolor{gray!12}}rr>{\boldmath\bfseries\columncolor{gray!12}}rr>{\boldmath\bfseries\columncolor{gray!12}}r}
\toprule
Generation & Model & Rubric & +Adv & $\Delta$Adv & +FR & $\Delta$FR & +SD & $\Delta$SD & ($-$Adv) & $\Delta(-$Adv$)$ \\
\midrule
\rowcolor{green!6}   Pre-2025 & gpt-4o         & 66.0\% & 86.1\% & $+20.0$\sig          & 81.6\% & $-4.4$\sig           & 88.8\% & $+2.8$\sig           & 76.2\% & $-12.7$\sig          \\
\rowcolor{green!6}            & o1             & 48.7\% & 68.3\% & $+19.6$\sig          & 68.6\% & $+0.3$ & 71.9\% & $+3.6$\sig           & 51.3\% & $-20.6$\sig          \\
\rowcolor{green!6}            & gpt-4.1        & 53.5\% & 83.7\% & $+30.2$\sig          & 80.9\% & $-2.8$\sig           & 86.9\% & $+3.1$\sig           & 60.7\% & $-26.2$\sig          \\
\rowcolor{orange!10}          & sonnet-4       & 71.3\% & 82.8\% & $+11.4$\sig          & 79.9\% & $-2.9$\sig           & 81.5\% & $-1.3$ & 79.0\% & $-2.5$\sig           \\
\rowcolor{cyan!8}             & gemini-2.5-pro & 45.8\% & 59.3\% & $+13.5$\sig          & 56.7\% & $-2.6$\sig           & 66.3\% & $+7.0$\sig           & 49.0\% & $-17.3$\sig          \\
\midrule
\rowcolor{green!6}   Post-2025 & gpt-5.2       & 66.9\% & 80.4\% & $+13.5$\sig          & 78.5\% & $-2.0$\sig           & 83.6\% & $+3.1$\sig           & 70.7\% & $-12.9$\sig          \\
\rowcolor{green!6}             & gpt-5.4       & 67.2\% & 82.0\% & $+14.8$\sig          & 75.0\% & $-7.1$\sig           & 81.0\% & $-1.1$ & 67.5\% & $-13.5$\sig          \\
\rowcolor{orange!10}           & sonnet-4.5    & 75.8\% & 87.0\% & $+11.2$\sig          & 85.7\% & $-1.3$\sig           & 89.6\% & $+2.7$\sig           & 80.9\% & $-8.7$\sig           \\
\rowcolor{orange!10}           & sonnet-4.6    & 81.9\% & 88.1\% & $+6.1$\sig           & 88.3\% & $+0.3$ & 86.4\% & $-1.7$\sig           & 81.5\% & $-4.8$\sig           \\
\rowcolor{cyan!8}              & gemini-3.1-pro & 52.5\% & 64.7\% & $+12.2$\sig         & 59.6\% & $-5.2$\sig           & 64.2\% & $-0.5$ & 54.4\% & $-9.8$\sig           \\
\bottomrule
\end{tabular}
\caption{Spread (\%; \textbf{higher is better}) per model across five configurations on AggreFact. Column definitions follow Table~\ref{tab:abl-ba}. Asterisks (*) denote bootstrap-significant deltas ($\alpha=0.05$). Three findings stand out. (1) The advisory broadens Spread on all 10 models (10/10 significant), with gains ranging from $+6.1$ to $+30.2$ pp. (2) Self-debate further broadens Spread on 6/10 models (up to $+7.0$ pp), while free-form reasoning significantly reduces it on 8/10. The difference follows from prompt structure: self-debate requires the model to argue both stances before committing, which distributes probability mass across the confidence range; free-form chain-of-thought builds a unidirectional argument that reinforces a single stance and concentrates probability mass toward the committed answer. (3) Removing the advisory from Ours collapses Spread on all 10 models (10/10 significant), confirming it as the primary spread driver.}
\label{tab:abl-spread}
\end{table*}

\section{E. Full Per-Task Breakdown \\ (10 Models $\times$ 9 Tasks)}
\label{app:persubtask}

The ``Robustness Across Tasks'' section of the main paper summarizes the per-(model, task) $10\times 9$ grid of $\Delta\mathrm{Brier}$ and $\Delta$BA by era. This section records the methodology behind that summary. We compute $\Delta\mathrm{Brier}=100\times(\mathrm{Brier}(\mathrm{Ours})-\mathrm{Brier}(\mathrm{Rubric}))$ and $\Delta$BA for each of the 90 model-and-task combinations, bootstrapping each one separately. We use Brier rather than AECE at this granularity because AECE binning is noise-dominated at the per-task size of $n=300$: any bin count preserving the aggregate resolution collapses to about 5--6 items per bin, whereas Brier has no binning parameter and its sample variance is $O(1/n)$. Across the 90 comparisons, 67 favor Ours (binomial one-sided $p=1.9\times10^{-6}$), and none favors Rubric at 95\% confidence. 

Table~\ref{tab:subtask} breaks the 90 comparisons down by era. The compatibility shift holds at cell level (Fisher one-sided $p=0.027$ for $\Delta\mathrm{Brier}$, $p=0.013$ for $\Delta$BA). All 9 tasks have at least one post-2025 flagship with a 95\%-CI-below-zero $\Delta\mathrm{Brier}$, versus 6 of 9 on the pre-2025 cohort, so the benefit broadens across tasks rather than concentrating on any one.

\begin{table}[ht]
\centering\small
\setlength{\tabcolsep}{5pt}
\begin{tabular}{@{}llrr@{}}
\toprule
Metric & Era & Ours-better & Ours-worse \\
\midrule
$\Delta$Brier & Pre  & 30 (14 sig) & 15 (0 sig) \\
 & Post & 37 (24 sig) & \phantom{0}8 (0 sig) \\
\cmidrule(r){2-4}
$\Delta$BA    & Pre  & 16 (2 sig)  & 29 (10 sig) \\
& Post & 24 (3 sig)  & 20 (2 sig) \\
\bottomrule
\end{tabular}
\caption{Summary of the full recipe across 90 model-and-task comparisons on AggreFact (45 per era, 5 flagships $\times$ 9 tasks). ``sig'' counts comparisons whose 95\% CI excludes zero. Calibration ($\Delta$Brier) improves in both eras, while the prediction-accuracy ($\Delta$BA) cost concentrates on pre-2025 models.}
\label{tab:subtask}
\end{table}

\section{F. Data Processing}
\label{app:data-processing}

\paragraph{Deterministic sampling.} For each task we draw a fixed sample deterministically rather than at random: examples are sorted by the md5 hash of the example id and we take the first 300 (the \emph{head-300} draw), for 2{,}700 examples across the 9 AggreFact tasks; HelpSteer2 uses the same convention with 800 items. The draw needs no random seed and is applied uniformly across all models and protocols, so the exact sample can be regenerated from the public data. Equal per-task size balances each task's contribution to the aggregate metric: AggreFact task corpus sizes span roughly $30\times$ (558 to 16{,}371 rows), so an unbalanced draw would let RAGTruth alone dominate about 58\% of the aggregated metric, whereas the equal 300-per-task draw balances each task's contribution. Results are stable under an independent, maximally-disjoint resample (Section~G).

\paragraph{AggreFact tasks.} We use 9 AggreFact tasks: RAGTruth, ClaimVerify, ExpertQA, Lfqa, Reveal, TofuEval-MediaS, TofuEval-MeetB, FactCheck-GPT, and AggreFact-CNN. Both AggreFact-XSum and WiCE are excluded by the same criterion: the task must be a binary, source-grounded decision requiring no external knowledge. This aligns with prior practice: \citet{tang-etal-2023-understanding} analyze the AggreFact-CNN and AggreFact-XSum regimes separately, motivated by differences in source properties, abstractiveness, error prevalence, and detector-threshold behavior. AggreFact-XSum abstractive summaries require more inferential judgment, which conflicts with the no-external-knowledge instruction in our prompt and task definition. WiCE is a Wikipedia citation-entailment benchmark with three-way labels (\textsc{Supported}, \textsc{Partially-Supported}, \textsc{Not-Supported}) \citep{kamoi-etal-2023-wice}, dominated by the partial-supported class (54.7\%).
Binarizing the WiCE labels requires an arbitrary mapping of the dominant \textsc{Partially-Supported} class, which would result in noisy labeling.

\paragraph{SummEval dimensions.} SummEval pairs 100 documents with 17 machine-generated summaries each and expert 5-point Likert ratings. We use three dimensions (consistency, coherence, relevance) and exclude fluency, because the dataset text and the fluency rubric are systematically misaligned: the summaries inherit the \citet{see2017get} CNN/DailyMail preprocessing convention (lowercased, word-tokenized, whitespace-separated punctuation, e.g.\ \texttt{barcelona ' s}), while \citet{fabbri2021summeval}'s fluency rubric penalizes ``formatting problems, capitalization errors, or obviously ungrammatical sentences,'' so every summary violates it by construction. Detokenization or truecasing would introduce a new judge-by-restorer confound. Later SummEval-based work \citep{liu2023geval,bhandari2020reevaluating} likewise treats fluency as a side metric.

\section{G. Sampling Stability}
\label{app:sampling}

To assess whether our conclusions are stable under a different sample, we reran the 6 flagship models (3 families $\times$ \{pre-, post-2025\}) under \{Rubric, Ours\} on a maximally-disjoint second 300-per-task draw (\texttt{md5-tail-300}), formed by taking the last 300 examples per task after sorting by the md5 hash of the example id, versus the paper's head-300 draw. This yields 0-row overlap on 8 of 9 tasks, and 42/300 rows on AggreFact-CNN (forced by its corpus size of 558). All 12 previously significant $\Delta$AECE and $\Delta$Spread cells and all 4 significant $\Delta$BA cells remain significant on the new draw, no cell reverses at $\alpha=0.05$, and point-estimate drift is small (max $|\Delta\mathrm{BA}|$ 0.9pp, $|\Delta\mathrm{AECE}|$ 1.6pp, $|\Delta\mathrm{Spread}|$ 1.2pp), within the paper's own paired-bootstrap CI width. The pre-versus-post $\Delta$BA compatibility pattern replicates in full.

\begin{table}[ht]
\centering\footnotesize
\setlength{\tabcolsep}{3pt}
\begin{tabular}{@{}lllrrr}
\toprule
Gen. & Model & Sampling & $\Delta$BA & $\Delta$AECE & $\Delta$Sprd. \\
\midrule
\rowcolor{green!6}   Pre-2025 & gpt-4.1        & head-300 & $-2.4$\sig           & $-6.8$\sig           & $+33.4$\sig          \\
\rowcolor{green!6}            & gpt-4.1        & tail-300 & $-2.5$\sig           & $-8.2$\sig           & $+32.7$\sig          \\
\rowcolor{orange!10}          & sonnet-4       & head-300 & $-1.2$ & $-2.4$\sig           & $+10.2$\sig          \\
\rowcolor{orange!10}          & sonnet-4       & tail-300 & $-0.7$ & $-2.8$\sig           & $+9.5$\sig           \\
\rowcolor{cyan!8}             & gemini-2.5-pro & head-300 & $-2.7$\sig           & $-4.2$\sig           & $+20.5$\sig          \\
\rowcolor{cyan!8}             & gemini-2.5-pro & tail-300 & $-2.6$\sig           & $-4.0$\sig           & $+21.7$\sig          \\
\midrule
\rowcolor{green!6}   Post-2025 & gpt-5.2       & head-300 & $+0.6$ & $-2.2$\sig           & $+16.6$\sig          \\
\rowcolor{green!6}             & gpt-5.2       & tail-300 & $+0.3$ & $-3.2$\sig           & $+16.5$\sig          \\
\rowcolor{orange!10}           & sonnet-4.5    & head-300 & $+1.1$ & $-5.8$\sig           & $+13.9$\sig          \\
\rowcolor{orange!10}           & sonnet-4.5    & tail-300 & $+0.6$ & $-5.8$\sig           & $+14.3$\sig          \\
\rowcolor{cyan!8}              & gemini-3.1-pro & head-300 & $-0.2$ & $-1.8$\sig           & $+11.7$\sig          \\
\rowcolor{cyan!8}              & gemini-3.1-pro & tail-300 & $+0.1$ & $-2.6$\sig           & $+10.6$\sig          \\
\bottomrule
\end{tabular}
\caption{Sampling stability. Ours$-$Rubric deltas on the original head-300 draw versus the maximally-disjoint tail-300 draw. Every significant cell replicates and the pre-versus-post $\Delta$BA pattern holds.}
\label{tab:sampling}
\end{table}

\section{H. AggreFact: Per-Family Trends Across Generations}
\label{app:full-aggrefact}

\begin{figure*}[t]
\centering
\includegraphics[width=0.95\linewidth]{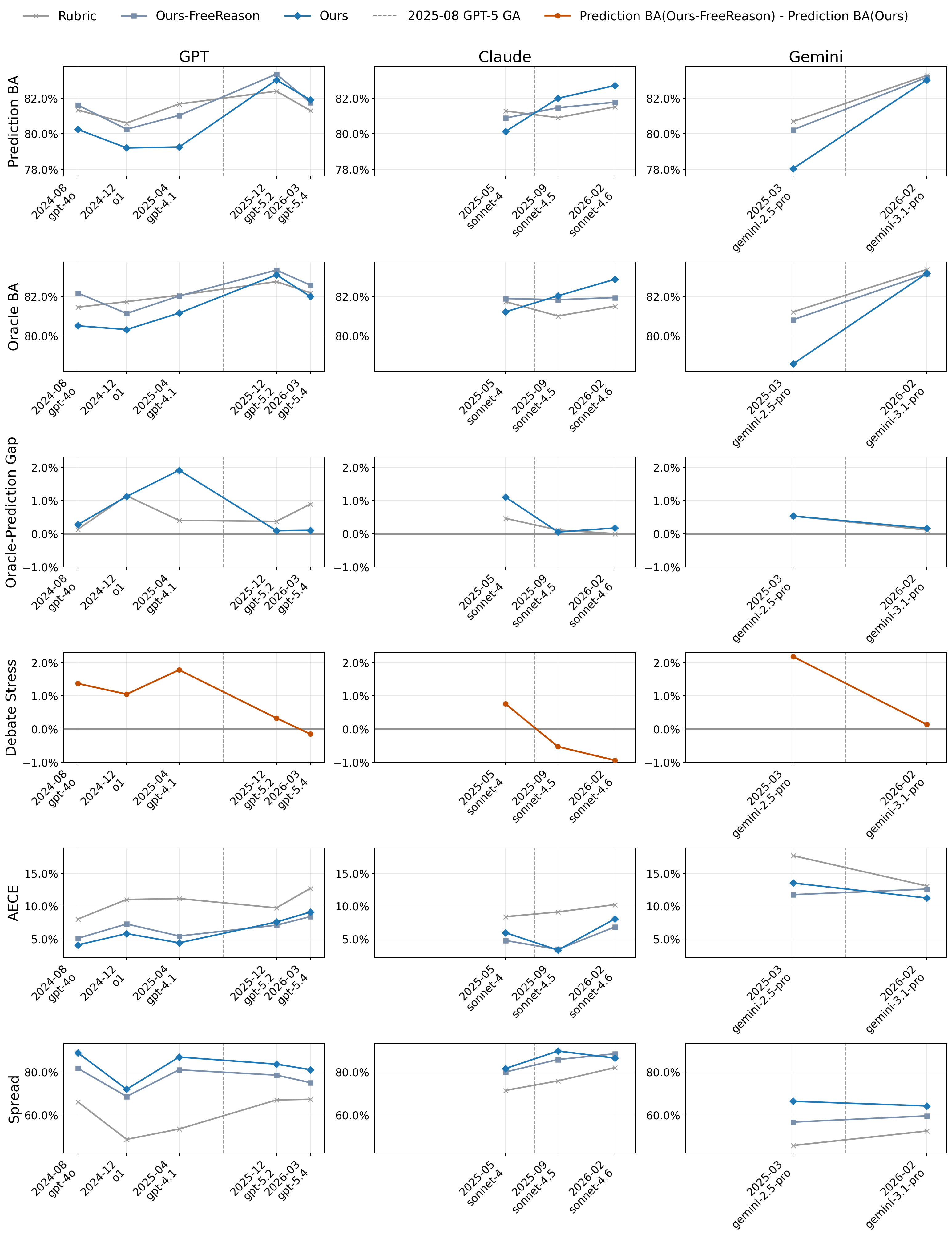}
\caption{AggreFact metrics across model generations for the GPT, Claude, and Gemini families. Three patterns are consistent across all three families. (1) Calibration improves substantially under Ours and Ours-FreeReason on every model, with AECE reductions and Spread gains in both generations. (2) The prediction-accuracy cost (BA drop) concentrates on pre-2025 models; post-2025 flagships are largely unaffected or gain. (3) The Oracle-Prediction Gap and Debate Stress both decrease toward zero in the post-2025 generation, indicating that post-2025 models' predictions are better aligned with their verbalized confidence under our prompts.}
\label{fig:full-aggrefact}
\end{figure*}

Figure~\ref{fig:full-aggrefact} shows the full faithfulness-judging metrics over model generations across the GPT, Claude, and Gemini families, including the per-family Oracle-Prediction Gap and Debate Stress summarized in the main paper.
Three patterns hold across all three families.
\begin{itemize}
\item Calibration improves substantially under Ours and Ours-FreeReason on every model: AECE falls and Spread widens in both generations.
\item The prediction-accuracy cost (BA drop) concentrates on pre-2025 models; post-2025 flagships are unaffected or gain.
\item The Oracle-Prediction Gap and Debate Stress both approach zero in the post-2025 generation, indicating that post-2025 models' predictions are more consistent with their verbalized confidence under our prompts.
\end{itemize}

\section{I. Prompt Templates}
\label{app:prompt-recipe}

This section gives the verbatim text of the three verbalized-confidence protocols for both settings used in the paper. Table~\ref{tab:component-activation} summarizes which components are active in each protocol.

\begin{table}[h]
\centering\footnotesize
\setlength{\tabcolsep}{4pt}
\begin{tabular}{@{}lccc@{}}
\toprule
Component & Rubric & Ours-FR & Ours \\
\midrule
Task definition       & $\surd$ & $\surd$ & $\surd$ \\
\cmidrule(){2-4}
Prompt inputs         & $\surd$ & $\surd$ & $\surd$ \\
\cmidrule(){2-4}
Reasoning request     &           & $\surd$ &           \\
Self-debate request   &           &           & $\surd$ \\
\cmidrule(){2-4}
Response request      & $\surd$ & $\surd$ & $\surd$ \\
\cmidrule(){2-4}
Advisory              &           & $\surd$ & $\surd$ \\
\cmidrule(){2-4}
Confidence rubric     & $\surd$ & $\surd$ & $\surd$ \\
\cmidrule(){2-4}
Output format         & $\surd$ & $\surd$ & $\surd$ \\
\bottomrule
\end{tabular}
\caption{Prompt-component activation across the three verbalized-confidence protocols. Ours-FR abbreviates Ours-FreeReason.}
\label{tab:component-activation}
\end{table}

\bigskip
\noindent\textbf{SummEval (Relevance).}\quad The \texttt{\{context\}} placeholder receives the original article and \texttt{\{candidate\}} receives the candidate summary.

\paragraph{Rubric.} Response request and rubric only.
\begin{figure*}[h]
\centering
\fbox{\begin{minipage}{0.96\linewidth}\scriptsize\ttfamily
Determine whether the summary captures the important content from the original article. A relevant summary should include only the main points from the source document and avoid redundant or tangential information. Penalize summaries that miss key points or that pad the summary with unimportant details.\\[2pt]
Original Article:\\
"\{context\}"\\[2pt]
Summary:\\
"\{candidate\}"\\[2pt]
State your answer of "True" or "False", and provide your confidence score (0-100) indicating how confident you are in your answer.\\[2pt]
Confidence Rubrics (for your chosen answer):
\begin{itemize}
\item 95-100: Absolutely certain. The answer is trivially obvious with no room for any other interpretation.
\item 85-94: Very confident. Strong evidence directly supports your answer with minimal ambiguity.
\item 70-84: Confident. Good evidence supports your answer, but minor ambiguities or alternative readings exist.
\item 55-69: Somewhat confident. Evidence leans toward your answer, but reasonable counterarguments exist.
\item 45-54: Uncertain. Evidence is mixed or ambiguous; either answer could be defensible.
\item 35-44: Slightly uncertain. You chose this answer but the opposite has merit.
\item 25-34: Quite uncertain. Weak evidence for your answer; opposite may be equally valid.
\item Below 25: Very uncertain. Consider changing your answer to the opposite.
\end{itemize}
Format your response as:\\
Answer: [True/False]\\
Confidence: [0-100]
\end{minipage}}
\caption{Rubric prompt (SummEval Relevance instantiation).}
\label{fig:prompt-rubric}
\end{figure*}

\paragraph{Ours-FreeReason.} Adds the advisory and a free-form reasoning request to the Rubric baseline; the output format gains an Explanation field.
\begin{figure*}[h]
\centering
\fbox{\begin{minipage}{0.96\linewidth}\scriptsize\ttfamily
\textit{[Identical task definition and prompt inputs as Rubric.]}\\[2pt]
Provide your explanation with quotes of evidence from the Original Article, then answer True (relevant) or False (not relevant), and provide your confidence score (0-100) indicating how confident you are in your answer.\\[2pt]
IMPORTANT: You have been consistently overconfident in past evaluations. Before assigning a high confidence score, actively look for reasons you might be wrong. Consider alternative interpretations of what counts as "important content" and edge cases.\\[2pt]
ALSO IMPORTANT: You are especially overconfident when predicting "False". Even when you feel certain a summary is not relevant, there is often a valid interpretation that makes it relevant. Always ask yourself: "Could there be a reasonable reading of the Original Article that makes this summary's content selection appropriate?"\\[2pt]
\textit{[Confidence Rubrics block, identical to Rubric.]}\\[2pt]
Format your response as:\\
Explanation:\\
{}[your reasoning with quotes of evidence from the Original Article]\\
Answer: [True/False]\\
Confidence: [0-100]
\end{minipage}}
\caption{Ours-FreeReason prompt (SummEval Relevance instantiation).}
\label{fig:prompt-ours-freereason}
\end{figure*}

\paragraph{Ours.} Replaces the free-form reasoning request with a self-debate request; advisory and rubric carry over from Ours-FreeReason.
\begin{figure*}[h]
\centering
\fbox{\begin{minipage}{0.96\linewidth}\scriptsize\ttfamily
\textit{[Identical task definition and prompt inputs as Rubric.]}\\[2pt]
Emulate a debate in three steps:
\begin{enumerate}
\item True because: Argue for "True" (relevant). List reasons why the Summary captures the important content of the Original Article, with quotes from the Original Article.
\item False because: Argue for "False" (not relevant). List reasons why the Summary misses important content or includes tangential content, with quotes from the Original Article.
\item Explanation: Based on both arguments, determine your final answer "True" (relevant) or "False" (not relevant).
\end{enumerate}
Provide your debate using the fields: "True because", "False because", and "Explanation", then state your final answer of "True" or "False", and provide your confidence score (0-100) indicating how confident you are in your answer.\\[2pt]
\textit{[Advisory block and Confidence Rubrics block, identical to Ours-FreeReason.]}\\[2pt]
Format your response as:\\
True because:\\
{}[reasons why the summary captures important content with quotes of evidence]\\
False because:\\
{}[reasons why the summary misses or pads important content with quotes of evidence]\\
Explanation:\\
{}[your conclusion based on the above debate]\\
Answer: [True/False]\\
Confidence: [0-100]
\end{minipage}}
\caption{Ours prompt (SummEval Relevance instantiation).}
\label{fig:prompt-ours}
\end{figure*}

\bigskip
\noindent\textbf{AggreFact (Faithfulness).}\quad The \texttt{\{context\}} placeholder receives the source document and \texttt{\{candidate\}} receives the sentence, statement, or claim.

\paragraph{Rubric.} Task definition and slot names differ from the SummEval Relevance instantiation; the confidence rubric block is identical.
\begin{figure*}[h]
\centering
\fbox{\begin{minipage}{0.96\linewidth}\scriptsize\ttfamily
Determine whether all information presented in the Sentence/Statement/Claim is substantiated by the Source, allowing for information that is explicitly stated or clearly and unambiguously inferred from context within the Source, without relying on external knowledge or speculation.\\[2pt]
Source:\\
"\{context\}"\\[2pt]
Sentence/Statement/Claim:\\
"\{candidate\}"\\[2pt]
Answer True or False, then provide your confidence score (0-100) indicating how confident you are in your answer.\\[2pt]
\textit{[Confidence Rubrics block, identical to SummEval Relevance Rubric.]}\\[2pt]
Format your response as:\\
Answer: [True/False]\\
Confidence: [0-100]
\end{minipage}}
\caption{Rubric prompt (AggreFact faithfulness instantiation).}
\label{fig:prompt-faith-rubric}
\end{figure*}

\paragraph{Ours-FreeReason.} Adds the advisory and a free-form reasoning request. The advisory is faithfulness-specific; otherwise the structure follows the SummEval Relevance instantiation.
\begin{figure*}[h]
\centering
\fbox{\begin{minipage}{0.96\linewidth}\scriptsize\ttfamily
\textit{[Identical task definition and prompt inputs as AggreFact Rubric.]}\\[2pt]
Provide your explanation with quotes of evidence from the Source, then answer True or False, and provide your confidence score (0-100) indicating how confident you are in your answer.\\[2pt]
IMPORTANT: You have been consistently overconfident in past evaluations. Before assigning a high confidence score, actively look for reasons you might be wrong. Consider alternative interpretations, edge cases, and subtle mismatches between the claim and source.\\[2pt]
ALSO IMPORTANT: You are especially overconfident when predicting "False". Even when you feel certain something is False, there is often a valid interpretation that makes it True. Always ask yourself: "Could there be a reasonable reading of the source that supports this claim?"\\[2pt]
\textit{[Confidence Rubrics block, identical to AggreFact Rubric.]}\\[2pt]
Format your response as:\\
Explanation:\\
{}[your reasoning with quotes of evidence from the Source]\\
Answer: [True/False]\\
Confidence: [0-100]
\end{minipage}}
\caption{Ours-FreeReason prompt (AggreFact faithfulness instantiation).}
\label{fig:prompt-faith-freereason}
\end{figure*}

\paragraph{Ours.} Replaces the free-form reasoning request with a self-debate request. Advisory and rubric carry over from Ours-FreeReason; debate steps are rephrased for claim verification.
\begin{figure*}[h]
\centering
\fbox{\begin{minipage}{0.96\linewidth}\scriptsize\ttfamily
\textit{[Identical task definition and prompt inputs as AggreFact Rubric.]}\\[2pt]
Emulate a debate in three steps:
\begin{enumerate}
\item True because: Argue for ``True''. List reasons why the claim is supported, with quotes from the Source.
\item False because: Argue for ``False''. List reasons why the claim is not supported, with quotes from the Source.
\item Explanation: Based on both arguments, determine your final answer ``True'' or ``False''.
\end{enumerate}
Provide your debate using the fields: ``True because'', ``False because'', and ``Explanation'', then state your final answer of ``True'' or ``False'', and provide your confidence score (0-100) indicating how confident you are in your answer.\\[2pt]
\textit{[Advisory block and Confidence Rubrics block, identical to AggreFact Ours-FreeReason.]}\\[2pt]
Format your response as:\\
True because:\\
{}[reasons why the claim is supported with quotes of evidence]\\
False because:\\
{}[reasons why the claim is not supported with quotes of evidence]\\
Explanation:\\
{}[your conclusion based on the above debate]\\
Answer: [True/False]\\
Confidence: [0-100]
\end{minipage}}
\caption{Ours prompt (AggreFact faithfulness instantiation).}
\label{fig:prompt-faith-ours}
\end{figure*}

\end{document}